\documentclass[runningheads]{llncs}
\usepackage{graphicx}
\usepackage{amsmath,amssymb} 
\usepackage{color}

\usepackage{gensymb}
\usepackage{booktabs}
\usepackage{float}
\usepackage{multirow}
\usepackage{subcaption}
\usepackage{algorithm}
\usepackage{algpseudocode}
\usepackage{hyperref}
\usepackage{mathtools}
\usepackage{wrapfig}
\usepackage{varwidth}
\usepackage{cite}

\usepackage[font=footnotesize,labelfont=bf]{caption}

\DeclareMathOperator*{\argmax}{arg\,max}

\begin{document}

\newcommand{\point}{
    \raise0.7ex\hbox{.}
    }


\pagestyle{headings}

\mainmatter

\title{Pano2Vid: Automatic Cinematography for Watching 360\degree~Videos} 

\titlerunning{Pano2Vid: Automatic Cinematography for Watching 360\degree~Videos} 

\authorrunning{Y.-C.~Su, D.~Jayaraman and K.~Grauman} 

\author{Yu-Chuan Su, Dinesh Jayaraman, and Kristen Grauman} 
\institute{The University of Texas at Austin} 

\maketitle

\begin{abstract}
We introduce the novel task of Pano2Vid --- \emph{automatic cinematography} in
panoramic 360\degree~videos.  Given a 360\degree~video, the goal is to direct
an imaginary camera to \emph{virtually} capture natural-looking normal
field-of-view (NFOV) video.  By selecting ``where to look" within the panorama
at each time step, Pano2Vid aims to free both the videographer and the end
viewer from the task of determining what to watch.  Towards this goal, we first
compile a dataset of 360\degree~videos downloaded from the web, together with
human-edited NFOV camera trajectories to facilitate evaluation.  Next, we
propose \textsc{AutoCam}, a data-driven approach to solve the Pano2Vid task.
\textsc{AutoCam}
leverages NFOV web video to discriminatively identify space-time ``glimpses" of
interest at each time instant, and then uses dynamic programming to select
optimal human-like camera trajectories.  Through experimental evaluation on
multiple newly defined Pano2Vid performance measures against several
baselines, we show that our method successfully produces informative videos
that could conceivably have been captured by human videographers.
\let\thefootnote\relax\footnotetext{Appears in Proceedings of Asian Conference on Computer Vision (ACCV'16)}
\end{abstract}

\section{Introduction}

\begin{figure}[t]
  \centering
  \includegraphics[width=1\textwidth]{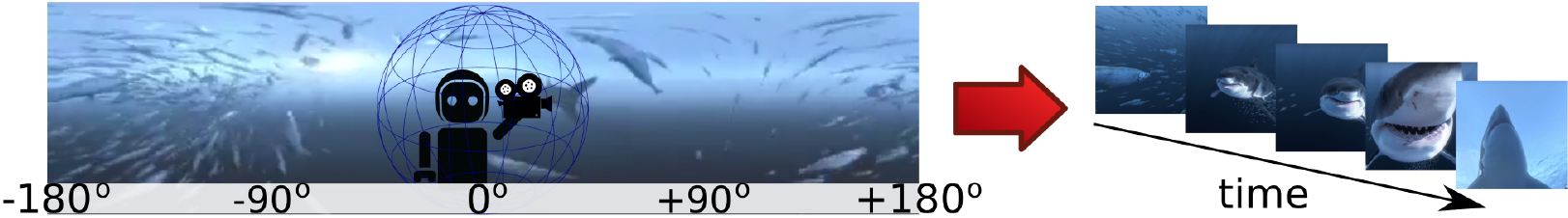}
  \vspace{-18pt}
  \caption{The ``Pano2Vid'' task: convert input 360\degree~video to output NFOV video.}
  \label{fig:conceptfig}
  \vspace{-15pt}
\end{figure}

A 360\degree~video camera captures the entire visual world as observable from
its optical center. This is a dramatic improvement over standard \emph{normal
field-of-view (NFOV)} video, which is usually limited to 65\degree. This
increase in field of view affords exciting new ways to record and experience
visual content.   For example, imagine a scientist in a shark cage studying the
behavior of a pack of sharks that are swimming in circles all around her. It
would be impossible for her to observe each shark closely in the moment. If,
however, she had a 360\degree~camera continuously recording spherical panoramic
video of the scene all around her, she could later replay her entire visual
experience hundreds of times, ``choosing her own adventure'' each time,
focusing on a different shark etc.  Similarly, a footballer wearing a
360\degree~camera could review a game from his in-game positions at all times,
studying passes which were open to him that he had not noticed in the heat of
the game.

In such cases and many others, 360\degree~video offers (1) a richer way to
experience the visual world unrestricted by a limited field of view, attention,
and cognitive capacity, even compared to actually being present \emph{in situ},
while (2) partially freeing the videographer of camera control.  Indeed,
360\degree~videos are growing increasingly popular as consumer- and
production-grade 360\degree~cameras (e.g., Ricoh, Bublcam, 360Fly, GoPro)
enter the market, and websites such as YouTube and Facebook begin to support
360\degree~content with viewers.

This new medium brings with it some new challenges.  Foremost, it largely
transfers the choice of ``where to look'' from the videographer to the viewer.
This makes 360\degree~video hard to view effectively, since a human viewer must
now somehow make the ``where to look'' choice and convey it to a video player
in real time.  Currently, there are three main approaches. In the first
approach, the user navigates a 360\degree~video manually.  A standard viewer
displays a small portion of the 360\degree~video corresponding to a normal
field-of-view (NFOV) camera\footnote{See, e.g.,
\url{https://www.youtube.com/watch?v=HNOT_feL27Y}, }.  The user must either
drag using a mouse, or click on up-down-left-right cursors, to adjust the
virtual camera pose continuously, for the full duration of the video.  A second
approach is to show the user the entire spherical panoramic video unwrapped
into its warped, equirectangular projection\footnote{See, e.g.,
\url{https://www.youtube.com/watch?v=Nv6MCkaR5mc}}.  While less effort for the
user, the distortions in this projected view make watching such video difficult
and unintuitive.  The third approach is to wear a virtual reality headset, a
natural way to view 360\degree~video that permits rich visual experiences.
However, a user must usually be standing and moving about, with a headset
obscuring all real-world visual input, for the full duration of the video. This
can be uncomfortable and/or impractical over long durations.  Plus, similar to
the click-based navigation, the user remains ``in the dark" about what is
happening elsewhere in the scene, and may find it difficult to decide where to
look to find interesting content in real time.  In short, all three
existing paradigms have interaction bottlenecks, and viewing 360\degree~video
remains cumbersome.

To address this difficulty, we define ``Pano2Vid'', a new computer vision
problem (see Fig~\ref{fig:conceptfig}). The task is to design an algorithm to
automatically control the pose and motion of a virtual NFOV camera within an
input 360\degree~video. The output of the system is the NFOV video captured by
this virtual camera. Camera control must be optimized to produce video that
could conceivably have been captured by a human observer equipped with a
\emph{real} NFOV camera. A successful Pano2Vid system would therefore take the
burden of choosing ``where to look'' off both the videographer and the end
viewer: the videographer could enjoy the moment without consciously
directing her camera, while the end viewer could watch intelligently-chosen
portions of the video in the familiar NFOV format.

For instance, imagine a Pano2Vid system that automatically outputs hundreds of
NFOV videos for the 360\degree~shark cage video, e.g., focusing on different
sharks/subgroups of sharks in turn. This would make analysis much easier for
the scientist, compared to manually selecting and watching hundreds of
different camera trajectories through the original 360\degree~video. A
machine-selected camera trajectory could also serve as a useful default
initialization for viewing 360\degree~content, where a user is not forced to
interact with the video player continuously, but could \emph{opt} to do so when
they desire. Such a system could even be useful as an editing aid for cinema.
360\degree~cameras could partially offload camera control from the
cinematographer to the editor, who might start by selecting from
machine-recommended camera trajectory proposals.

This work both formulates the Pano2Vid problem and introduces the first
approach to address it.  The proposed ``\textsc{AutoCam}" approach first learns
a discriminative model of human-captured NFOV web video. It then uses this
model to identify candidate viewpoints and events of interest to capture in
360\degree~video, before finally stitching them together through optimal camera
motions using a dynamic programming formulation for presentation to human
viewers.  Unlike prior attempts at automatic cinematography, which focus on
virtual 3D worlds and employ heuristics to encode popular idioms from
cinematography~\cite{christianson1996declarative,he1996virtual,elson2007lightweight,mindek2015automatized,foote2000flycam,sun2005region},
\textsc{AutoCam} is (a) the first to tackle real video from dynamic cameras and
(b) the first to consider directly \emph{learning} cinematographic tendencies
from data.

The contributions of this work are four-fold: (1) we formulate the computer
vision problem of automatic cinematography in 360\degree~video (Pano2Vid), (2)
we propose a novel Pano2Vid algorithm (\textsc{AutoCam}), (3) we compile a
dataset of 360\degree~web video, annotated with ground truth human-directed
NFOV camera
trajectories\footnote{\url{http://vision.cs.utexas.edu/projects/Pano2Vid}\label{url}}~and
(4) we propose a comprehensive suite of objective and subjective evaluation
protocols to benchmark Pano2Vid task performance. We benchmark \textsc{AutoCam}
against several baselines and show that it is the most successful at virtually
capturing natural-looking NFOV video.

\vspace{-8pt}

\section{Related Work}

\vspace{-5pt}

\subsubsection{Video summarization} 
Video summarization methods condense videos in \emph{time} by identifying
important events~\cite{truong2007video}.   A summary can take the form of a
keyframe
sequence~\cite{goldman2006schematic,yongjae2012discovering,kim2013jointly,khosla2013large,xiong2014detecting,gong2014diverse},
a sequence of video highlight
clips~\cite{gygli2014creating,gygli2015video,potapov2014category,zhao2014quasi,sun2014ranking,song2015tvsum},
or montages of frames~\cite{sun2014salient} or video clip
excerpts~\cite{pritch2007webcam}.   Among these, our proposed \textsc{AutoCam}
shares
with~\cite{kim2013jointly,khosla2013large,xiong2014detecting,sun2014ranking,song2015tvsum}
the idea of using user-generated visual content from the web as exemplars for
informative content.   However, whereas existing methods address
\emph{temporal} summarization of NFOV video, we consider a novel form of
\emph{spatial} summarization of 360\degree~video.  While existing methods
decide \emph{which frames to keep} to shorten a video, our problem is instead
to choose \emph{where to look} at each time instant.  Moreover, existing
summarization work assumes video captured intentionally by a camera person (or,
at least, a well-placed surveillance camera).  In contrast, our input videos
largely lack this deliberate control.  Moreover, we aim not only to capture
all \emph{important} content in the original 360\degree~video, but to do so in
a \emph{natural, human-like} way so that the final output video resembles video
shot by human videographers with standard NFOV cameras.

\vspace{-0.15in}
\subsubsection{Camera selection for multi-video summarization} 
Some efforts address \emph{multi-video}
summarization~\cite{fu2010multi,dale2012multi,arev2014automatic}, where the
objective is to select, at each time instant, video feed from one camera among
many to include in a summary video.   The input cameras are human-directed,
whether stationary or dynamic~\cite{arev2014automatic}.  In contrast, we deal
with a single hand-held 360\degree~camera, which is not intentionally directed
to point anywhere.

\vspace{-0.19in}
\subsubsection{Video retargeting}
Video retargeting aims to adapt a video to better suit the aspect ratio of a
target display, with minimal loss of content that has already been purposefully
selected by an editor or
videographer~\cite{liu2006video,avidan2007seam,rubinstein2008improved,krahenbuhl2009system,khoenkaw2015automatic}.
In our setting, 360\degree~video is captured \emph{without} human-directed
content selection; instead, the system must automatically select the content to
capture.  Furthermore, the spatial downsampling demanded by Pano2Vid will
typically be much greater than that required in retargeting.

\vspace{-0.19in}
\subsubsection{Visual saliency}
Salient regions are usually defined as those that capture the visual attention
of a human observer, e.g., as measured by gaze tracking. While saliency
detectors most often deal with static
images~\cite{harel2006graph,liu2011learning,achanta2009frequency,perazzi2012saliency},
some are developed for video
\cite{itti2005bayesian,zhai2006visual,guo2010novel,rudoy2013learning,wang2016learning},
including work that models temporal continuity in
saliency~\cite{rudoy2013learning}. Whereas saliency methods aim to capture
where human eyes move subconsciously during a free-viewing task, our Pano2Vid
task is instead to capture where human videographers would \emph{consciously
point their cameras}, for the specific purpose of capturing a video that is
\emph{presentable to other human viewers}. In our experiments, we empirically
verify that saliency is not adequate for automatic cinematography.

\vspace{-0.19in}
\subsubsection{Virtual cinematography}
Ours is the first attempt to automate cinematography in complex real-world
settings. Existing virtual cinematography work focuses on camera manipulation
within much simpler \emph{virtual} environments/video
games~\cite{christianson1996declarative,he1996virtual,elson2007lightweight,mindek2015automatized},
where the perception problem is bypassed~(3-D positions and poses of all
entities are knowable, sometimes even controllable), and there is full freedom
to position and manipulate the camera. Some prior
work~\cite{foote2000flycam,sun2005region} attempts virtual camera control
within restricted \emph{static} wide field-of-view video of classroom and video
conference settings, by tracking the centroid of optical flow in the scene. In
contrast, we deal with unrestricted 360\degree~web video of complex real-world
scenes, captured by moving amateur videographers with shaky hand-held devices,
where such simple heuristics are insufficient.  Importantly, our approach is
also the first to \emph{learn content-based camera control from data}, rather
than relying on hand-crafted guidelines/heuristics as all prior attempts do.

\vspace{-6pt}
\section{Approach}

\vspace{-4pt}
We first define the Pano2Vid problem in more detail (Sec.~\ref{sec:probdef})
and describe our data collection process (Sec.~\ref{sec:data}).  Then we
introduce our \textsc{AutoCam} approach (Sec.~\ref{sec:algo}).  Finally, we introduce
several evaluation methodologies for quantifying performance on this complex
task (Sec.~\ref{sec:metrics}), including an annotation collection procedure to
gather human-edited videos for evaluation (Sec.~\ref{sec:humanedit-collect}).

\vspace{-4pt}

\subsection{Pano2Vid Problem Definition}\label{sec:probdef}

First, we define the Pano2Vid task of automatic videography for
360\degree~videos. Given a dynamic panoramic 360\degree~video, the goal is to
produce ``natural-looking'' normal-field-of-view (NFOV) video. For this work,
we define NFOV as spanning a horizontal angle of 65.5\degree~(corresponding to
a typical 28mm focal length full-frame Single Lens Reflex Camera~\cite{SUN360})
with a 4:3 aspect ratio.

Broadly, a natural-looking NFOV video is one which is indistinguishable
from human-captured NFOV video (henceforth ``HumanCam''). Our ideal video
output should be such that it could conceivably have been captured by a human
videographer equipped with an NFOV camera whose optical center coincides
exactly with that of the 360\degree~video camera, with the objective of best
presenting the event(s) in the scene.  In this work, we do not allow skips in
time nor camera zoom, so the NFOV video is defined completely by the camera
trajectory, i.e., the time sequence of the camera's principal axis directions. To
solve the Pano2Vid problem, a system must determine a NFOV camera trajectory
through the 360\degree~video to carve it into a HumanCam-like NFOV video.

\vspace{-6pt}

\subsection{Data Collection: 360\degree~Test Videos and NFOV Training Videos}\label{sec:data}

\vspace{-4pt}

Human-directed camera trajectories are content-based and often present scenes
in \emph{idiomatic} ways that are specific to the situations, and with specific
intentions such as to tell a story~\cite{mascelli1998five}.  Rather than
hand-code such characteristics through cinematographic
rules/heuristics~\cite{christianson1996declarative,he1996virtual,elson2007lightweight,mindek2015automatized},
we propose to \emph{learn} to capture NFOV videos, by observing HumanCam videos
from the web.  The following overviews our data collection procedure.

\vspace{-12pt}

\subsubsection{$360\degree$ videos}
We collect $360\degree$ videos from YouTube using the keywords ``Soccer,''
``Mountain Climbing,'' ``Parade,'' and ``Hiking.'' These terms were selected to
have (i) a large number of relevant 360\degree~video results, (ii) dynamic
activity, i.e., spatio-temporal \emph{events}, rather than just static scenes,
and (iii) possibly multiple regions/events of interest at the same time.  For
each query term, we download the top 100 videos sorted by relevance and filter
out any that are not truly $360\degree$ videos (e.g., animations, slide shows
of panoramas, restricted FOV) or have poor lighting, resolution, or stitching
quality.  This yields a Pano2Vid test set of 86 total $360\degree$ videos with
a combined length of $7.3$ hours.  
See the project webpage\textsuperscript{\ref{url}} for example videos.

\vspace{-12pt}

\subsubsection{HumanCam NFOV videos}

In both the learning stage of \textsc{AutoCam} (Sec~\ref{sec:algo}) and the proposed
evaluation methods (Sec~\ref{sec:metrics}), we need a model for HumanCam.  We
collect a large diverse set of HumanCam NFOV videos from YouTube using the same
query terms as above and imposing a per-video max length of 4 minutes.  For
each query term, we collect about 2,000 videos, yielding a final HumanCam set
of 9,171 videos totalling 343 hours. See Sec.~\ref{sec:appendix_youtube} for details.

\vspace{-4pt}

\subsection{AutoCam: Proposed Solution for the Pano2Vid Task}\label{sec:algo}

We now present \textsc{AutoCam}, our approach to solve the Pano2Vid task. The
input to the system is an arbitrary 360\degree~video, and the output is a
natural looking NFOV video extracted from it.

\textsc{AutoCam} works in two steps. First, it evaluates all virtual NFOV
spatio-temporal ``glimpses" (ST-glimpses) sampled from the $360\degree$ video
for their ``capture-worthiness''---their likelihood of appearing in HumanCam
NFOV video. Next, it selects virtual NFOV camera trajectories, prioritizing
both (i) high-scoring ST-glimpses from the first step, and (ii) smooth
human-like camera movements.  \textsc{AutoCam} is fully automatic and does not
require any human input. Furthermore, as we will see next, the proposed
learning approach is unsupervised---it learns a model of human-captured NFOV
video simply by watching clips people upload to YouTube.

\begin{figure*}[t]
  \centering
  \vspace{8pt}
  \begin{subfigure}[b]{0.66\textwidth}
    \begin{minipage}{0.5\textwidth}
      \vspace{-24pt}
      \includegraphics[width=\textwidth]{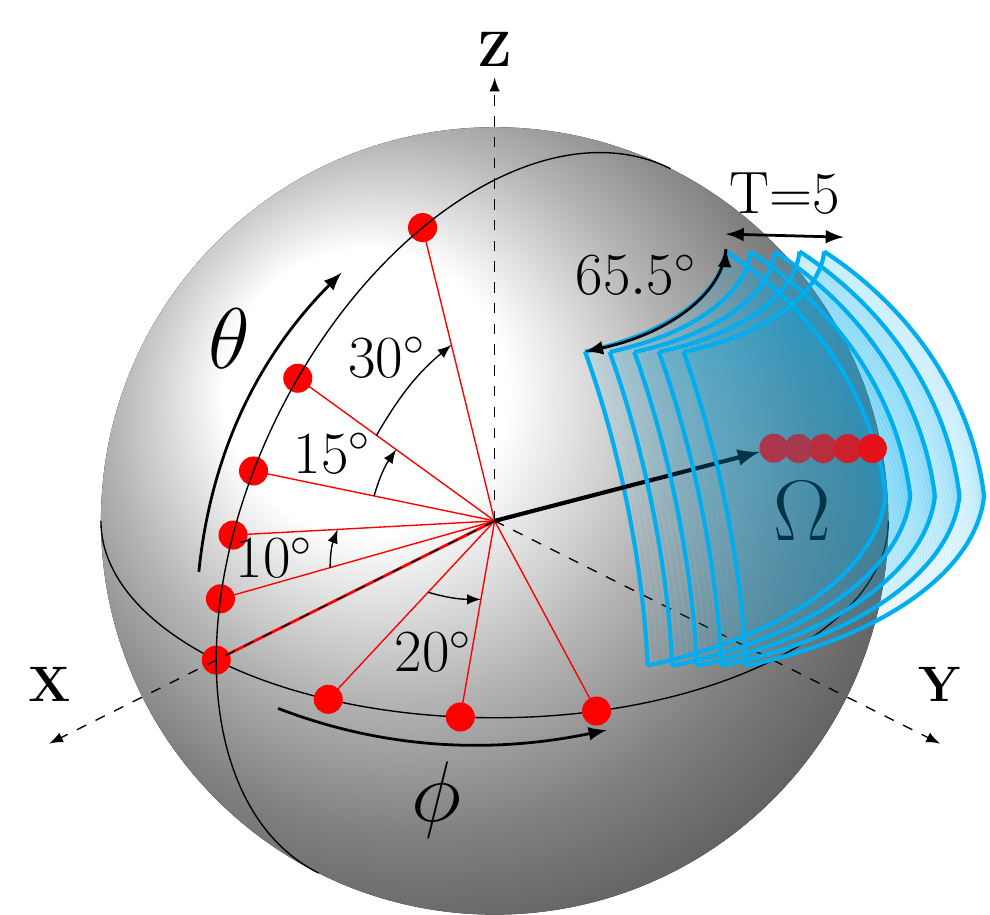}
    \end{minipage}
    \begin{minipage}{0.48\textwidth}
      \includegraphics[width=\textwidth]{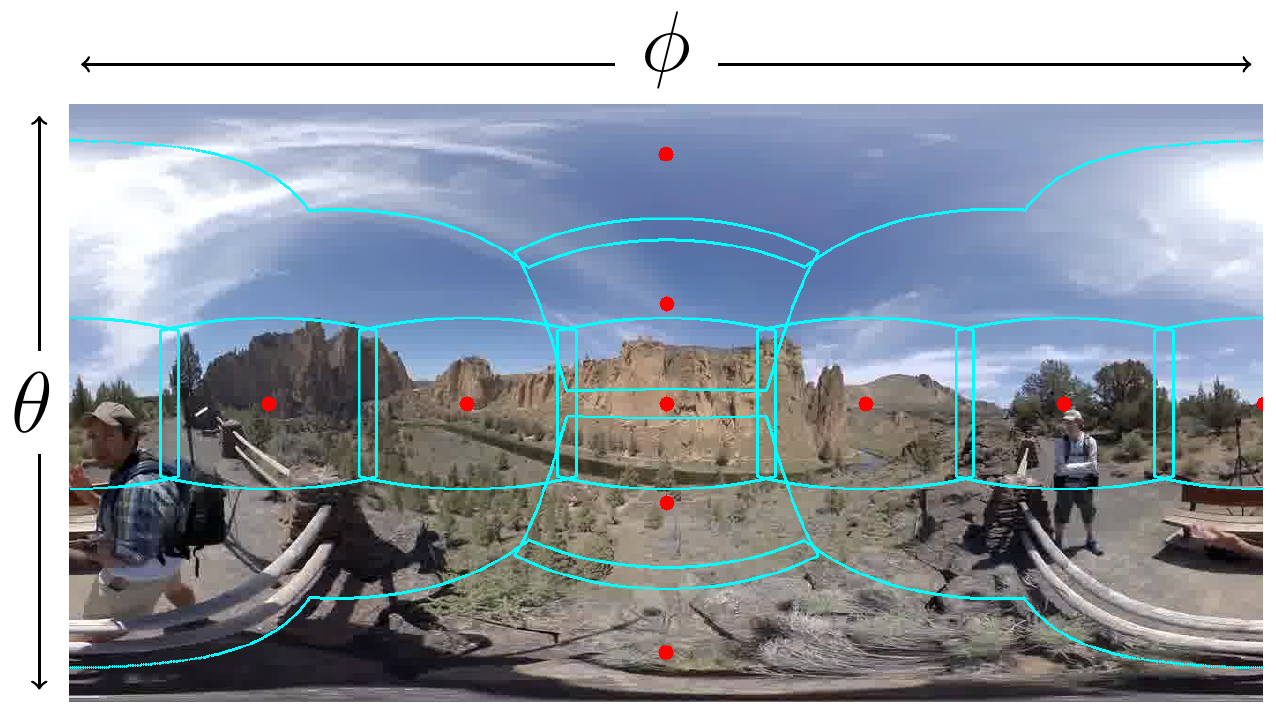}
    \end{minipage}
    \vspace{-4pt}
    \subcaption{Sample ST-glimpses and score capture-worthiness.}
    \label{fig:sample_glimpse}
  \end{subfigure}
  \begin{subfigure}[b]{0.33\textwidth}
    \includegraphics[width=\textwidth]{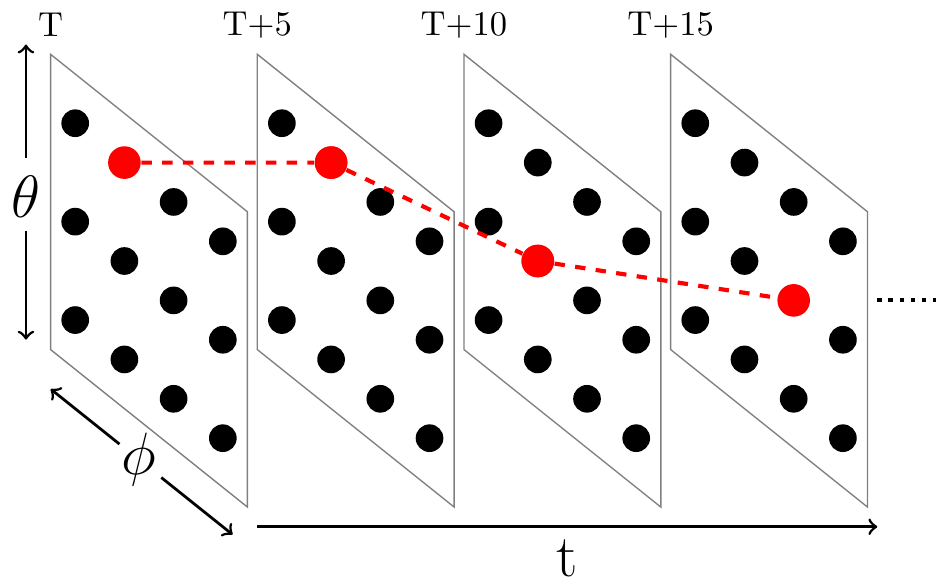}
    \vspace{-4pt}
    \subcaption{Stitch glimpses.}
    \label{fig:dp_stitching}
  \end{subfigure}
  \vspace{-18pt}
  \caption{\textsc{AutoCam} first samples and scores the capture-worthiness of
ST-glimpses. It then jointly selects a glimpse for each time step and stitches them together to form the output NFOV video.  Best viewed in color.}
  \label{fig:approach}
  \vspace{-12pt}
\end{figure*}

\vspace{-15pt}

\subsubsection{Capture-worthiness of spatio-temporal glimpses}

The first stage aims to find content that is likely to be captured by human
videographers. We achieve this by scoring the capture-worthiness of
candidate ST-glimpses sampled from the $360\degree$ video.  An ST-glimpse is a
five-second NFOV video clip recorded from the $360\degree$ video by directing
the camera to a fixed direction in the $360\degree$ camera axes. One such glimpse
is depicted as the blue stack of frame excerpts on the surface of the
sphere in Fig~\ref{fig:sample_glimpse}. These are not rectangular regions in
the equirectangular projection (Fig~\ref{fig:sample_glimpse}, right) so they
are projected into NFOV videos before processing.   We sample candidate
ST-glimpses at longitudes $\phi \in \Phi = \{0,20,40,\ldots,340\}$ and
latitudes $\theta \in \Theta = \{0,\pm10,\pm20,\pm30,\pm45,\pm75\}$ and
intervals of 5 seconds. Each candidate ST-glimpse is defined by the camera
principal axis direction $(\theta,\phi)$
and time $t$: $\Omega_{t, \theta, \phi}
\equiv (\theta_t,\phi_t) \in \Theta \times \Phi$.
See Sec.~\ref{sec:appendix_sampling}.

Our approach learns to score capture-worthiness from HumanCam data.  We expect
capture-worthiness to rely on two main facets: content and composition.  The
\emph{content} captured by human videographers is naturally very diverse. For
example, in a mountain climbing video, people may consider capturing the
recorder and his companion as well as a beautiful scene such as the sunrise as
being equally important. Similarly, in a soccer video, a player dribbling and a
goalkeeper blocking the ball may both be capture-worthy.   Our approach
accounts for this diversity both by learning from a wide array of NFOV HumanCam
clips and by targeting related domains via the keyword query data collection
described above.  The \emph{composition} in HumanCam data is a meta-cue,
largely independent of semantic content, that involves the framing effects
chosen by a human videographer.  For example, an ST-glimpse that captures only
the bottom half of a human face is not capture-worthy, while a framing that
captures the full face is; a composition for outdoor scenes may tend to put
the horizon towards the middle of the frame, etc.

Rather than attempt to characterize capture-worthiness through rules,
\textsc{AutoCam} \emph{learns} a data-driven model.  We make the following
hypotheses: (i) the majority of content in HumanCam NFOV videos were considered
capture-worthy by their respective videographers (ii) most random ST-glimpses
would \emph{not} be capture-worthy. Based on these hypotheses, we train a
capture-worthiness classifier. Specifically, we divide each HumanCam video into
non-overlapping 5-second clips, to be used as positives, following (i) above.
Next, \emph{all} candidate ST-glimpses extracted from (disjoint) $360\degree$
videos are treated as negatives, per hypothesis (ii) above. Due to the weak
nature of this supervision, both positives and negatives may have some label
noise.

\begin{figure}[t]
  \centering
  \includegraphics[width=\textwidth]{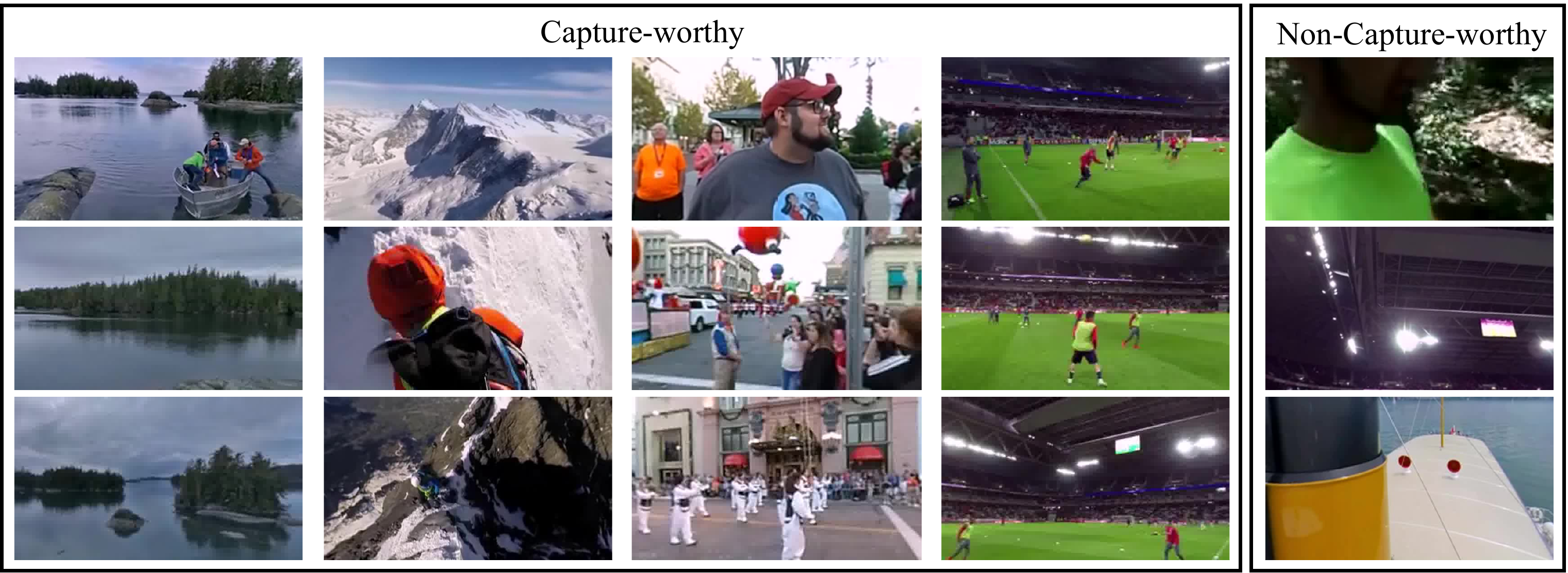}
  \vspace{-15pt}
  \caption{Example glimpses scored by \textsc{AutoCam}. Left 4 columns are glimpses considered
  capture-worthy by our method; each column is from the same time step in the
  same video. Right column shows non-capture-worthy glimpses.  }
  \label{fig:glimpses_example}
  \vspace{-18pt}
\end{figure}

To represent each ST-glimpse and each 5s HumanCam clip, we use off-the-shelf
convolutional 3D features (C3D)~\cite{tran2015learning}. C3D is a generic video
feature based on 3D (spatial+temporal) convolution that captures appearance and
motion information in a single vector representation, and is known to be useful
for recognition tasks.  We use a leave-one-video-out protocol to train one
capture-worthiness classifier for each $360\degree$ video. Both the positive
and negative training samples are from videos returned by the same keyword
query term as the test video, and we sub-sample the $360\degree$ videos so that
the total number of negatives is twice that of positives. We use logistic
regression classifiers; positive class probability estimates of ST-glimpses
from the left-out video are now treated as their capture-worthiness scores.

Fig~\ref{fig:glimpses_example} shows examples of ``capture-worthy'' and
``non-capture-worthy'' glimpses as predicted by our system.  We see that there
may be multiple capture-worthy glimpses at the same moment, and both the
content and composition are important for capture-worthiness.  Please see
the Sec.~\ref{sec:appendix_naturalness} for further analysis, including a study of
how our predictions correlate with the viewpoint angles.

\subsubsection{Camera trajectory selection}
\label{sec:trajectory_stitching}

\vspace{-12pt}

After obtaining the capture-worthiness score of each candidate ST-glimpse, we
construct a camera trajectory by finding a path over the ST-glimpses that
maximizes the \emph{aggregate} capture-worthiness score, while simultaneously
producing human-like smooth camera motions. A naive solution would be to choose
the glimpse with the maximum score at each step. This trajectory would capture
the maximum aggregate capture-worthiness, but the resultant NFOV video may have
large/shaky unnatural camera motions. For example, when two ST-glimpses located
in opposite directions on the viewing sphere have high capture-worthiness
scores, such a naive solution would end up switching between these two
directions at every time-step, producing unpleasant and even physically
impossible camera movements.

\begin{wrapfigure}{L}{0.65\textwidth}
\vspace{-1.5cm}
\begin{minipage}{0.66\textwidth}
\begin{algorithm}[H]
  \caption{Camera trajectory selection}
  \label{alg:stitch}
  \begin{algorithmic}
    \State $C~\gets$ Capture-worthiness scores
    \State $\epsilon \gets$ Valid camera motion
    \ForAll{$\theta, \phi$}
      \State $Accum[\Omega_{1, \theta, \phi}]~\gets~C[\Omega_{1, \theta, \phi}]$
    \EndFor
    \For{$t \gets 2, T$}
        \ForAll{$\theta, \phi$}
            \State \begin{varwidth}[t]{\linewidth}
            $\Omega_{t-1, \theta^{\prime}, \phi^{\prime}}~\gets~\argmax_{\theta^{\prime}, \phi^{\prime}}
            Accum[\Omega_{t-1, \theta^{\prime}, \phi^{\prime}}]$ \par
            \hskip2.2cm $s.t.~|\Omega_{t, \theta, \phi} - \Omega_{t-1, \theta^{\prime}, \phi^{\prime}}| \le \epsilon$
            \end{varwidth}
            \State $Accum[\Omega_{t, \theta, \phi}]~\gets~Accum[\Omega_{t-1, \theta^{\prime}, \phi^{\prime}}]+C[\Omega_{t, \theta, \phi}]$
            \State $TraceBack[\Omega_{t, \theta, \phi}]~\gets~\Omega_{t-1, \theta^{\prime}, \phi^{\prime}}$
      \EndFor
    \EndFor
    \State $\Omega~\gets~\argmax_{\theta, \phi} Accum[\Omega_{T, \theta, \phi}]$
    \For{$t \gets T,1$}
      \State $Traj[t]~\gets~\Omega$
      \State $\Omega~\gets~TraceBack[\Omega]$
    \EndFor
  \end{algorithmic}
\end{algorithm}
\end{minipage}
\vspace{-0.7cm}
\end{wrapfigure}

Instead, to construct a trajectory with more human-like camera operation, we
introduce a \emph{smooth motion} prior when selecting the ST-glimpse at each
time step.  Our prior prefers trajectories that are stable over those that jump
abruptly between directions. For the example described above, the smooth prior
would suppress trajectories that switch between the two directions constantly
and promote those that focus on one direction for a longer amount of time. In
practice, we realize the smooth motion prior by restricting the trajectory from
choosing an ST-glimpse that is displaced from the previous ST-glimpse by more
than $\epsilon= 30\degree$ in both longitude and latitude, i.e.
\begin{equation}
\setlength{\abovedisplayskip}{4pt}
\setlength{\belowdisplayskip}{2pt}
  \label{eq:allowed_motion}
  |\Delta \Omega|_{\theta} = |\theta_{t}-\theta_{t-1}| \le \epsilon, \;
  |\Delta \Omega|_{\phi} = |\phi_{t}-\phi_{t-1}| \le \epsilon.
\end{equation}

Given (i) the capture-worthiness scores of all candidate ST-glimpses and (ii)
the smooth motion constraint for trajectories, the problem of finding the
trajectories with maximum aggregate capture-worthiness scores can be reduced to
a shortest path problem. Let $C(\Omega_{t,\theta,\phi})$ be the
capture-worthiness score of the ST-glimpse at time $t$ and viewpoint
$(\theta,\phi)$.  We construct a 2D lattice per time slice, where each node
corresponds to an ST-glimpse at a given angle pair.  The edges in the lattice
connect ST-glimpses from time step $t$ to $t+1$, and the weight for an edge is
defined by:
\begin{equation}
\setlength{\abovedisplayskip}{4pt}
\setlength{\belowdisplayskip}{4pt}
  \label{eq:edge_weight}
  E\left(\Omega_{t,\theta,\phi}, \Omega_{t+1,\theta^\prime,\phi^\prime}\right) =
  \begin{cases}
    -C(\Omega_{t+1,\theta^\prime,\phi^\prime}), & |\Omega_{t,\theta,\phi}-\Omega_{t+1,\theta^\prime,\phi^\prime}| \le \epsilon \; \\
    \infty, & \text{otherwise},
  \end{cases}
\end{equation}
where the difference above is shorthand for the two angle requirements in
Eq.~\ref{eq:allowed_motion}.  See Fig~\ref{fig:dp_stitching}, middle and right.

The solution to the shortest path problem over this graph now corresponds to
camera trajectories with maximum aggregate capture-worthiness. This solution
can be efficiently computed using dynamic programming. See pseudocode in
Alg~\ref{alg:stitch}.  At this point, the optimal trajectory indicated by this
solution is ``discrete'' in the sense that it makes jumps between discrete
directions after each 5-second time-step.  To smooth over these jumps, we
linearly interpolate the trajectories between the discrete time instants, so
that the final camera motion trajectories output by \textsc{AutoCam} are
continuous. In practice, we generate $K$ NFOV outputs from each $360\degree$
input by (i) computing the best trajectory ending at each ST-glimpse location
(of 198 possible), and (ii) picking the top $K$ of these.

Note that \textsc{AutoCam} is an offline batch processing algorithm that
watches the entire video before positioning the virtual NFOV camera at each
frame.  This matches the target setting of a human editing a pre-recorded
$360\degree$ video to capture a virtual NFOV video, as the human is free to
watch the video in full.   In fact, we use human-selected edits to help
evaluate \textsc{AutoCam}
(Sec~\ref{sec:humanedit-collect},~\ref{sec:humanedit-eval}).

\vspace{-6pt}
\subsection{Quantitative Evaluation of Pano2Vid Methods}\label{sec:metrics}

Next we present evaluation metrics for the Pano2Vid problem.  A good metric
must measure how close a Pano2Vid algorithm's output videos are to
human-generated NFOV video, while simultaneously being reproducible for easy
benchmarking in future work. We devise two main criteria:
\begin{itemize}
\vspace{-4pt}
  \item \textbf{HumanCam-based metrics}: Algorithm outputs should look like
      HumanCam videos---the more indistinguishable the algorithm outputs are
      from real manually captured NFOV videos, the better the algorithm.
  \item \textbf{HumanEdit-based metrics}: Algorithms should select camera
    trajectories close to human-selected trajectories (``HumanEdit'')---The
    closer algorithmically selected camera motions are to those selected by
    humans editing the same $360\degree$ video, the better the algorithm.
\vspace{-4pt}
\end{itemize}
The following fleshes out a family of metrics capturing these two criteria.
All of which can easily be reproduced and compared to easily, given the same
training/testing protocol is applied.

\vspace{-6pt}

\subsubsection{HumanCam-based metrics}

We devise three HumanCam-based metrics:

\vspace{-6pt}

\paragraph{\textbf{Distinguishability}: Is it possible to distinguish Pano2Vid and HumanCam outputs?}

Our first metric quantifies \emph{distinguishability} between algorithmically
generated and HumanCam videos. For a fully successful Pano2Vid algorithm, these
sets would be entirely indistinguishable. This method can be considered as an
automatic Turing test that is based on feature statistics instead of human
perception; it is also motivated by the adversarial network
framework~\cite{goodfellow2014generative} where the objective of the generative
model is to disguise the discriminative model.  We measure distinguishability
using 5-fold cross validation performance of a discriminative classifier
trained with HumanCam videos as positives, and algorithmically generated videos
as negatives.  Training and testing negatives in each split are generated from
disjoint sets of $360\degree$ video.

\vspace{-6pt}
\paragraph{\textbf{HumanCam-likeness}: Which Pano2Vid method gets closer to HumanCam?}
\label{sec:distance_to_positive_set}

This metric directly compares outputs of multiple Pano2Vid methods using their
relative distances from HumanCam videos in a semantic feature space (e.g., C3D
space). Once again a classifier is trained on HumanCam videos as positives, but
this time with \emph{all} algorithm-generated videos as negatives. Similar to
exemplar SVM~\cite{malisiewicz-iccv11}, each algorithm-generated video is
assigned a ranking based on its distance from the decision boundary
(i.e.~HumanCam-likeness), using a leave-one-360\degree-video-out training and
testing scheme. We rank all Pano2Vid algorithms for each $360\degree$ video and
compare their normalized mean rank; lower is better.  We use
classification score rather than raw feature distance because we are only
interested in the factors that distinguish Pano2Vid and HumanCam.
Since this metric depends on the relative comparison of all methods, it
requires the output of all methods to be available during evaluation.

\vspace{-6pt}
\paragraph{\textbf{Transferability}: Do semantic classifiers transfer between Pano2Vid and HumanCam video?}

This metric tries to answer the question: if we learn to distinguish between
the 4 classes (based on search keywords) on HumanCam videos, would the
classifier perform well on Pano2Vid videos (Human $\rightarrow$ Auto), and vice
versa (Auto $\rightarrow$ Human)?   Intuitively, the more similar the domains,
the better the transferability. A similar method is used to evaluate automatic
image colorization in~\cite{zhang2016colorful}. To quantify transferability, we
train a multi-class classifier on Auto(/Human) videos generated by a given
Pano2Vid method and test it on Human(/Auto) videos. This test accuracy is the
method's transferability score.

\vspace{-6pt}
\subsubsection{HumanEdit-based metrics}

Our metrics thus far compare Pano2Vid outputs with generic NFOV videos.  We now
devise a set of HumanEdit-based metrics that compare algorithm outputs to
human-selected NFOV camera trajectories, given the \emph{same input
360\degree~video}. Sec~\ref{sec:humanedit-collect} will explain how we obtain
HumanEdit trajectories. Note that a single 360\degree~video may have several
equally valid HumanEdit camera trajectory annotations, e.g. from different
annotators. 

\vspace{-6pt}
\paragraph{\textbf{Mean cosine similarity}: How closely do the camera trajectories match?}

To compute this metric, we first measure the frame-wise cosine distance (in the
$360\degree$ camera axes) between the virtual NFOV camera principal axes selected
by Pano2Vid and HumanEdit. These frame-wise distances are then pooled into one
score in two different ways:  (1) \textbf{Trajectory pooling}: Each Pano2Vid
trajectory is compared to its best-matched HumanEdit trajectory. Frame-wise
cosine distances to each human trajectory are first averaged. Each Pano2Vid
output is then assigned a score corresponding to the minimum of its average
distance to HumanEdit trajectories. Trajectory pooling rewards Pano2Vid outputs
that are similar to at least one HumanEdit trajectory \emph{over the whole
video}, and (2) \textbf{Frame pooling}: This pooling method rewards Pano2Vid
outputs that are similar to different HumanEdit tracks in different portions of
the video. First, each frame is assigned a score based on its minimum
frame-wise cosine distance to a HumanEdit trajectory. Now, we simply average
this over all frames to produce the ``frame distance'' score for that
trajectory.  Frame pooling rewards Pano2Vid outputs that are similar to any
HumanEdit trajectory at each frame.

\vspace{-6pt}

\paragraph{\textbf{Mean overlap}: How much do the fields of view overlap?}

The cosine distance between principal axes ignores the fact that cameras have
limited FOV. To account for this, we compute ``overlap'' metrics on Pano2Vid
and HumanEdit camera FOVs on the unit sphere. Specifically, we approximate the
overlap using $\max(1-\frac{\Delta\Omega}{\text{FOV}},0)$, which is $1$ when
the pricipal axes coincide, and $0$ for all $\Delta \Omega > \text{FOV}$. We
apply both trajectory and frame pooling as for the cosine distance.

\vspace{-4pt}

\subsection{HumanEdit Annotation Collection}\label{sec:humanedit-collect}

\begin{figure*}[t]
  \vspace{-8pt}
  \centering
  \includegraphics[width=0.8\textwidth]{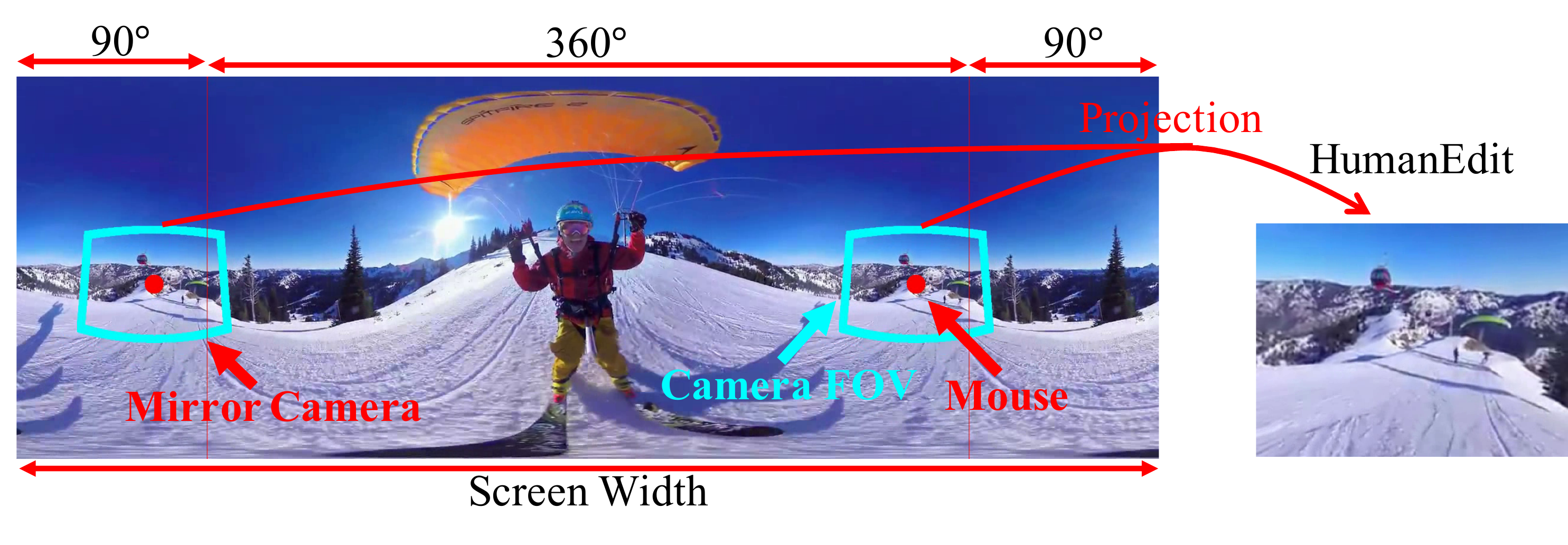}
  \vspace{-12pt}
  \caption{HumanEdit interface. We display the 360\degree~video in
  equirectangular projection and ask annotators to direct the camera using the
  mouse. The NFOV video is rendered and displayed to the annotator offline.  Best viewed in color.}
  \label{fig:humancam_interface}
  \vspace{-12pt}
\end{figure*}

To collect human editors' annotations, we ask multiple annotators to watch the
$360\degree$ test videos and generate the camera trajectories from them.  We
next describe the annotation collection process. We then analyze the
consistency of collected HumanEdit trajectories.

\vspace{-4pt}

\subsubsection{Annotation interface and collection process}

Fig~\ref{fig:humancam_interface} shows the HumanEdit annotation interface. We
display the entire 360\degree~video in equirectangular projection.  Annotators
are instructed to move a cursor to direct a virtual NFOV camera.  Virtual NFOV
frame boundaries are backprojected onto the display (shown in cyan) in real
time as the camera is moved. 

We design the interface to mitigate problems due to discontinuities at the
edges. First, we extend the panoramic strip by $90\degree$ on the left and
right as shown in Fig~\ref{fig:humancam_interface}. The cursor may now smoothly
move over the $360\degree$ boundaries to mimic camera motion in the real world.
Second, when passing over these boundaries, content is duplicated, and so is
the cursor position and frame boundary rendering. When passing over an edge of
this extended strip, the cursor is repositioned to the duplicated position that
is already on-screen by this time. Finally, before each annotation, our
interface allows the annotator to pan the panoramic strip to a chosen longitude
to position important content centrally if they so choose. Please refer to
Sec.~\ref{sec:appendix_interface} for more visual examples and project webpage for the interface in action.

For each $360\degree$ video, annotators watch the full video first to
familiarize themselves with its content. Thus, they have the same information
as our \textsc{AutoCam} approach. Each annotator provides \emph{two} camera trajectories
per 360\degree~video, to account for the possibility of multiple good
trajectories. Each of 20 $360\degree$ videos is annotated by 3 annotators,
resulting in a final database with 120 human-annotated trajectories adding up
to 202 minutes of NFOV video.  Our annotators were 8 students aged between
24--30.

\subsubsection{HumanEdit consistency analysis}

After collecting HumanEdit, we measure the consistency between trajectories
annotated by different annotators using the metrics described in
Sec~\ref{sec:metrics}.

Table~\ref{tab:annotation_analysis} shows the results. The average cosine
distance between human trajectories is $0.520$, which translates to $59\degree$
difference in camera direction at every moment. The difference is significant,
considering the NFOV is $65\degree$.  Frame differences, however, are much
smaller---$37\degree$ on average, and overlap of $>50\%$ across annotators at
every frame.  These differences indicate that there is more than one natural
trajectory for each $360\degree$ video, and different annotators may pick
different trajectories.  Still, with $>50\%$ overlap at any given moment, we
see that there is often something in the $360\degree$ video that catches
everyone's eye; different trajectories arise because people place different
priority on the content and choose to navigate through them in different
manner.  Overall, this analysis justifies our design to ask each annotator to
annotate twice and underscores the need for metrics that take the multiple
trajectories into account, as we propose.

\begin{table*}[t]
  \vspace{-8pt}
  \begin{minipage}{0.42\textwidth}\scriptsize
      \centering
      \caption{HumanEdit consistency.}
      \label{tab:annotation_analysis}
      \setlength{\tabcolsep}{4pt}
      \begin{tabular}{llc}
        \toprule
        Cosine     & Trajectory & 0.520\\
        Similarity & Frame      & 0.803\\
        \midrule
        \multirow{2}{*}{Overlap}& Trajectory & 0.462 \\
                                & Frame      & 0.643\\
        \bottomrule
      \end{tabular}
  \end{minipage}
  \begin{minipage}{0.58\textwidth}
    \centering
    \begin{subfigure}[t]{0.4\textwidth}
      \centering
      \includegraphics[width=\textwidth]{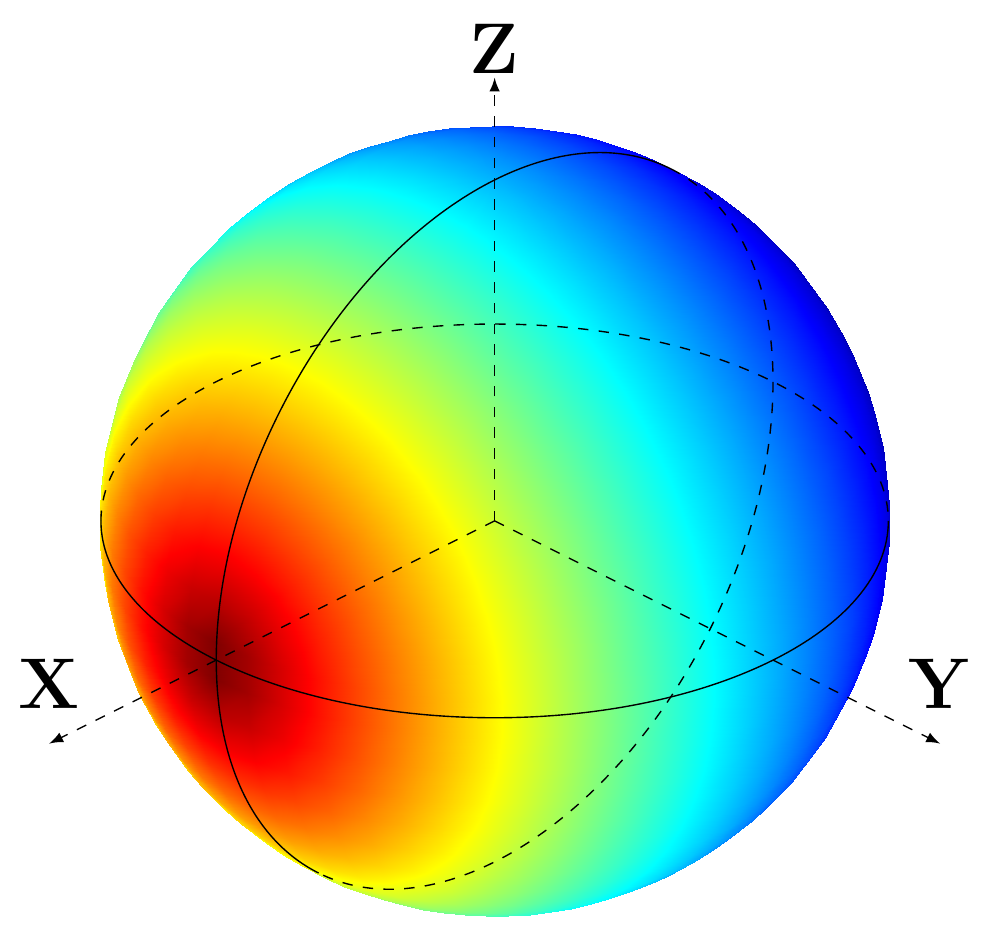}%
      \vspace{-8pt}
      \caption{\textsc{Center}}%
      \label{fig:center_prior}%
    \end{subfigure}
    ~
    \begin{subfigure}[t]{0.4\textwidth}
      \centering
      \includegraphics[width=\textwidth]{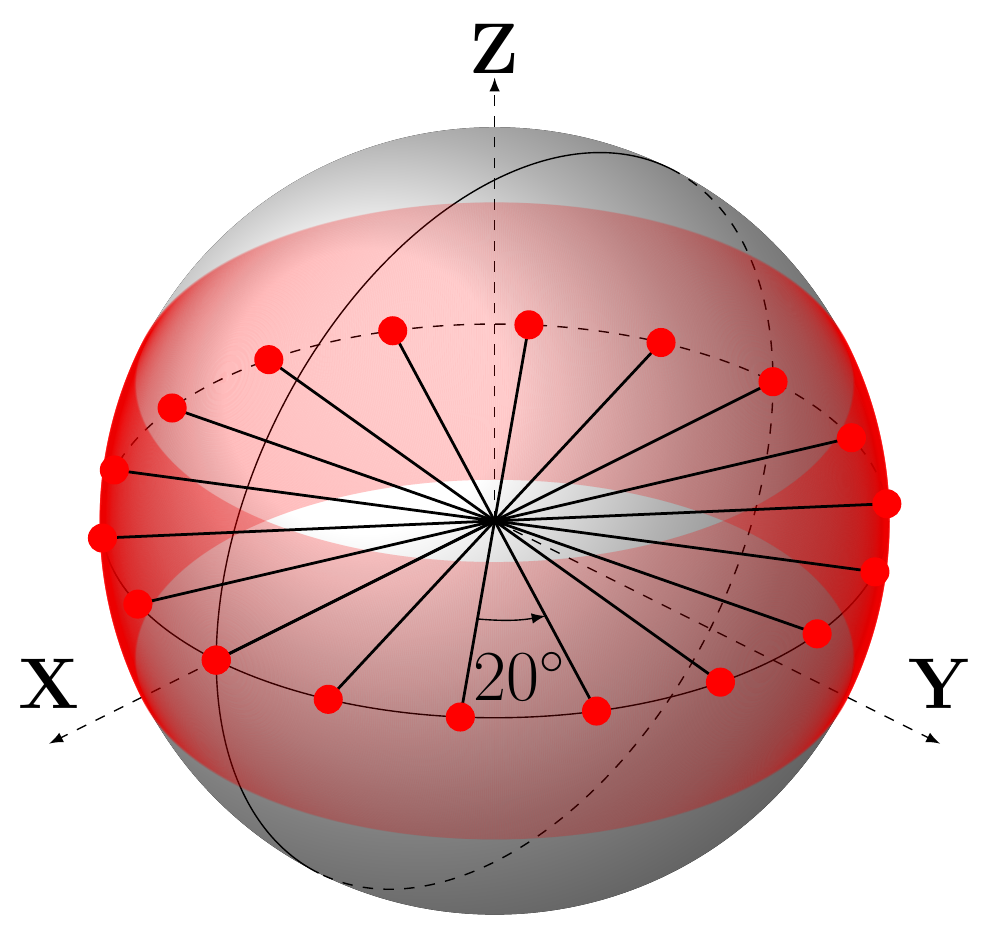}%
      \vspace{-8pt}
      \caption{\textsc{Eye-level}}%
      \label{fig:eyelevel_prior}%
    \end{subfigure}
    \vspace{-8pt}
    \caption{Baseline illustration. \textsc{Center} generates random
        trajectories biased toward the $360\degree$ video center.
    \textsc{Eye-level} generates static trajectories on the equator.}
    \label{fig:baselines}
  \end{minipage}
  \vspace{-24pt}
\end{table*}

\vspace{-4pt}

\section{Experiment}\label{sec:results}

\vspace{-4pt}

\subsubsection{Baseline}

We compare our method to the following baselines.

\begin{itemize}
\item \textsc{Center} $+$ Smooth motion prior---This baseline biases camera
    trajectories to lie close to the ``center'' of the standard
    360\degree~video axes, accounting for the fact that user-generated
    360\degree~videos often have regions of interest close to the centers of
    the standard equirectangular projection.  We sample from a Gaussian
    starting at the center, then centered at the current direction for
    subsequent time-steps.  See Table 2a.
\item \textsc{Eye-level} prior--- This baseline points the NFOV camera to some
    pre-selected direction on the equator (i.e.~at $0\degree$ latitude)
    throughout the video. $0\degree$ latitude regions often correspond to
    eye-level in our dataset. We sample at $20\degree$ longitudinal intervals
    along the equator. See Table 2b.
\item \textsc{Saliency} trajectory---This baseline uses our pipeline, but
    replaces our discriminative capture-worthiness scores with the saliency
    score from a popular saliency method~\cite{harel2006graph}.
\item \textsc{AutoCam w/o stitching}---This is an ablated variant of our method
    that has no camera motion constraint. We generate multiple outputs by
    sampling ST-glimpses at each time step based on their capture-worthiness
    scores.
\end{itemize}
For each method we generate $K$=20 outputs per $360\degree$
video, and average their results.  See Sec.~\ref{sec:appendix_multitraj} for details.

\vspace{-6pt}

\subsubsection{Implementation Details}

Following~\cite{tran2015learning}, we split the input video into 16 frame clips
then extract the $fc6$ activation for each clip and average them as the C3D video
features (whether on glimpses or full HumanCam clips). We use temporally
non-overlapping clips to reduce computation cost.  We use the Sport1M model
provided by the authors without fine-tuning.  We use logistic regression with
$C=1$ in all experiments involving a discriminative classifier.

\vspace{-6pt}

\subsection{HumanCam-based Evaluation}
\label{sec:experiment_youtube}
\begin{table}[t]\scriptsize
    \vspace{-8pt}
    \centering
    \setlength{\tabcolsep}{4pt}
    \caption{Pano2Vid performance: HumanCam-based metrics. The arrows in
    column 3 indicate whether lower scores are better ($\Downarrow$), or higher scores ($\Uparrow$).}
    \label{tab:human_cam_yt}
    \resizebox{\textwidth}{!}{
    \begin{tabular}{llcccc|cc}
        & & & \multirow{2}{*}{\textsc{Center}} &
        \multirow{2}{*}{\textsc{Eye-level}} &
        \multirow{2}{*}{\textsc{Saliency}} & \textsc{AutoCam w/o} &
        \textsc{AutoCam}\\
        & & & & & & \textsc{stitching} (ours)& (ours)\\
    \hline
        Distinguishability & Error rate (\%) & $\Uparrow$ & 1.3 & 2.8 & 5.2 & 4.0 & \textbf{7.0}\\ 
    \hline
        HumanCam-Likeness  & Mean Rank & $\Downarrow$ & 0.659 & 0.571 & 0.505 & 0.410 & \textbf{0.388} \\
    \hline
        \multirow{2}{*}{Transferability} & Human $\rightarrow$ Auto &
        \multirow{2}{*}{$\Uparrow$} & 0.574 & 0.609 & 0.595 & \textbf{0.637} & 0.523 \\
                                         & Auto $\rightarrow$ Human & & 0.526 & 0.559 & 0.550 & 0.561 & \textbf{0.588} \\
    \noalign{\hrule height .8pt}
    \end{tabular}
}
  \vspace{-4pt}
\end{table}

We first evaluate our method using the HumanCam-based metrics (defined in
Sec~\ref{sec:metrics}).  Table~\ref{tab:human_cam_yt} shows the results. Our
full method (\textsc{AutoCam}) performs the best in nearly all metrics.
It improves the Distinguishability and HumanCam-Likeness by $35\%$
and $23\%$ respectively, compared to the best baseline. The advantage in
Transferability is not as significant, but is still $5\%$ better in
Auto $\rightarrow$  Human transfer.
\textsc{AutoCam w/o stitching} is second-best overall, better than all other 3
baselines. These results establish that both components of our method---(i)
capture-worthy ST-glimpse selection, and (ii) smooth camera motion
selection---capture important aspects of human-like NFOV videos.

\begin{figure*}[t]
  \centering
  \hspace{-24pt}
  \includegraphics[width=1.05\textwidth]{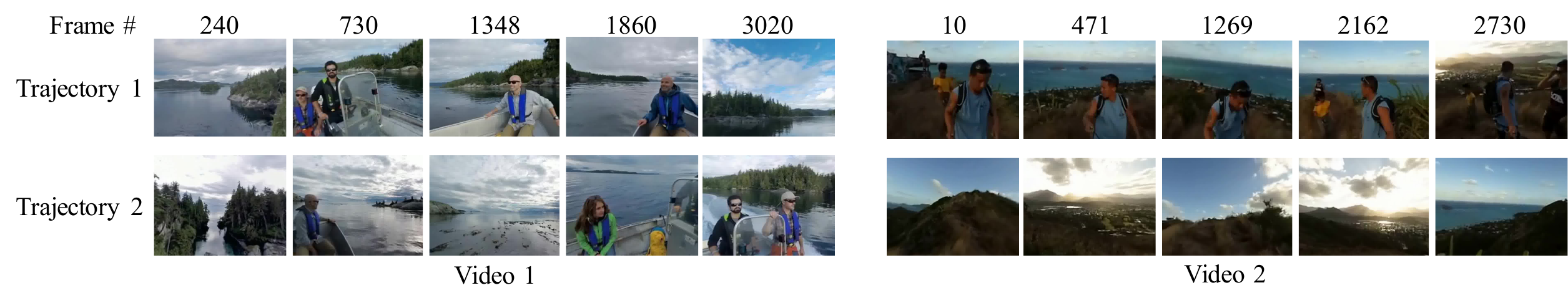}
  \vspace{-12pt}
  \caption{Example \textsc{AutoCam} outputs. We show the result for two $360\degree$
  videos, and two trajectories for each.
  }
  \label{fig:trajectory_example}
  \vspace{-12pt}
\end{figure*}

Among the remaining baselines, \textsc{Saliency}, which is content-based and
also uses our smooth motion selection pipeline, performs significantly better
than \textsc{Center} and \textsc{Eye-level}, which are uniformly poor
throughout. However, \textsc{Saliency} falls well short of even \textsc{AutoCam
w/o stitching}, establishing that saliency is a poor proxy for
capture-worthiness.

We observe that the transferability metric results are asymmetric, and
\textsc{AutoCam} only does best on transferring semantic classifiers in the
Auto $\rightarrow$ Human direction.  Interestingly, \textsc{AutoCam w/o
stitching} is best on Human $\rightarrow$ Auto, but the smooth motion
constraint adversely affects this score for \textsc{AutoCam}. This performance
drop may be caused by the content introduced when trying to stitch two
spatially disjoint capture-worthy ST-glimpses. While \textsc{AutoCam w/o
stitching} can jump directly between such glimpses, \textsc{AutoCam} is
constrained to move through less capture-worthy content connecting them.
Moreover, intuitively, scene recognition relies more on content selection than
on camera motion, so incoherent motion might not disadvantage \textsc{AutoCam
w/o stitching}.

Fig~\ref{fig:trajectory_example} shows example output NFOV videos of our
algorithm for two $360\degree$ videos.  For each video, we show two different
generated trajectories. Our method is able to find multiple natural NFOV videos
from each input.
See project webpage for video examples and comparisons of different methods.

\vspace{-6pt}

\subsection{HumanEdit-based Evaluation}\label{sec:humanedit-eval}

\begin{table*}[t]\scriptsize
    \vspace{-8pt}
    \centering
    \setlength{\tabcolsep}{4pt}
    \caption{Pano2Vid performance: HumanEdit-based metrics. Higher is better.}
    \label{tab:human_cam_human}
    \begin{tabular}{clccc|cc}
        & & \multirow{2}{*}{\textsc{Center}} &
        \multirow{2}{*}{\textsc{Eye-level}} &
        \multirow{2}{*}{\textsc{Saliency}} & \textsc{AutoCam w/o} &
        {\textsc{AutoCam}}\\
        & & & & & \textsc{stitching} (ours) & (ours) \\
        \hline
        Cosine     & Trajectory               & 0.257 & 0.268 & 0.063 & 0.184 & \textbf{0.304} \\
        Similarity & Frame                    & 0.572 & 0.575 & 0.387 & 0.541 & \textbf{0.581} \\
        \hline
        \multirow{2}{*}{Overlap} & Trajectory & 0.194 & 0.243 & 0.094 & 0.202 & \textbf{0.255} \\
        & Frame      & 0.336 & \textbf{0.392} & 0.188 & 0.354 & 0.389 \\
    \noalign{\hrule height .8pt}
    \end{tabular}
    \vspace{-12pt}
\end{table*}

Next, we evaluate all methods using the HumanEdit-based metrics
(Sec~\ref{sec:metrics}). Table~\ref{tab:human_cam_human} shows the results.
Once again, our method performs best on all but the frame-pooling overlap metrics.

On the cosine distance metric in particular, \textsc{AutoCam w/o stitching}
suffers significantly from having incoherent camera motion.  \textsc{Eye-level}
is second best on these metrics. It does better on frame-wise metrics,
suggesting that humans rarely choose static eye-level trajectories. Further,
\textsc{Eye-level} does better on overlap metrics, even outperforms
\textsc{AutoCam} on average per-frame overlap, suggesting a tendency to make
large mistakes which are penalized by cosine metrics but not by overlap
metrics.  \textsc{Saliency} scores poorly throughout; even though saliency
may do well at predicting human gaze fixations, as discussed above, this is not
equivalent to predicting  plausible NFOV excerpts.

To sum up, our method performs consistently strongly across a wide range of
metrics based on both resemblance to generic YouTube NFOV videos, and on
closeness to human-created edits of 360\degree~video. This serves as strong
evidence that our approach succeeds in capturing human-like virtual NFOV videos.

\vspace{-8pt}

\section{Conclusion}

\vspace{-4pt}

We formulate Pano2Vid: a new computer vision problem that aims to produce a
natural-looking NFOV video from a dynamic panoramic $360\degree$ video. We
collect a new dataset for the task, with an accompanying suite of Pano2Vid
performance metrics. We further propose \textsc{AutoCam}, an approach to learn
to generate camera trajectories from human-generated web video.  We hope that
this work will provide the foundation for a new line of research that requires
both scene understanding and active decision making. In the future, we plan to
explore supervised approaches to leverage HumanEdit data for learning the
properties of good camera trajectories and incorporate more task specific
features such as human detector.

\vspace{3mm}
\noindent {\bf Acknowledgement}.
This research is supported in part by NSF IIS -1514118 and a gift from Intel.
We also gratefully acknowledge the support of Texas Advanced Computing Center
(TACC).

\bibliographystyle{splncs}
\bibliography{0327}


\clearpage
\appendix

\textbf{Please see \url{http://vision.cs.utexas.edu/projects/Pano2Vid} for example AutoCam output,
HumanEdit annotation, and the annotation interface.}

\vspace{-8pt}

\section{HumanCam Videos Collection}
\label{sec:appendix_youtube}

\vspace{-8pt}

We use the Youtube api with the ``videoDuration'' filter set to short
(i.e.~\textless 4 minutes,) to search for NFOV videos. Because the Youtube api
only returns a limited number (about 500) of results for each query, we perform
4 queries for each term with the ``publish time'' filter set to \{2016, 2015,
2014, None\} respectively.  Duplicate results are removed (by Youtube ID).

\vspace{-8pt}
\section{ST Glimpses Sampling}
\label{sec:appendix_sampling}

\vspace{-8pt}

We perform dense sampling for ST glimpses both spatially and temporally.
Spatial samples are drawn from spherical coordinates following the camera
model~\cite{SUN360}. The samples are chosen to be as dense as possible to
enable fine camera control, but sparse enough to meet computation resource
constraints. We sample time and longitude $\phi$ uniformly, but more densely
around the equator for latitude $\theta$, because adjacent glimpses with the same
polar angle will have larger overlap at high latitude.

\vspace{-8pt}
\section{Capture-worthiness Analysis}
\label{sec:appendix_naturalness}

\vspace{-8pt}

\begin{figure*}[t]
  \centering
    \begin{subfigure}[t]{0.32\textwidth}
        \includegraphics[width=\textwidth]{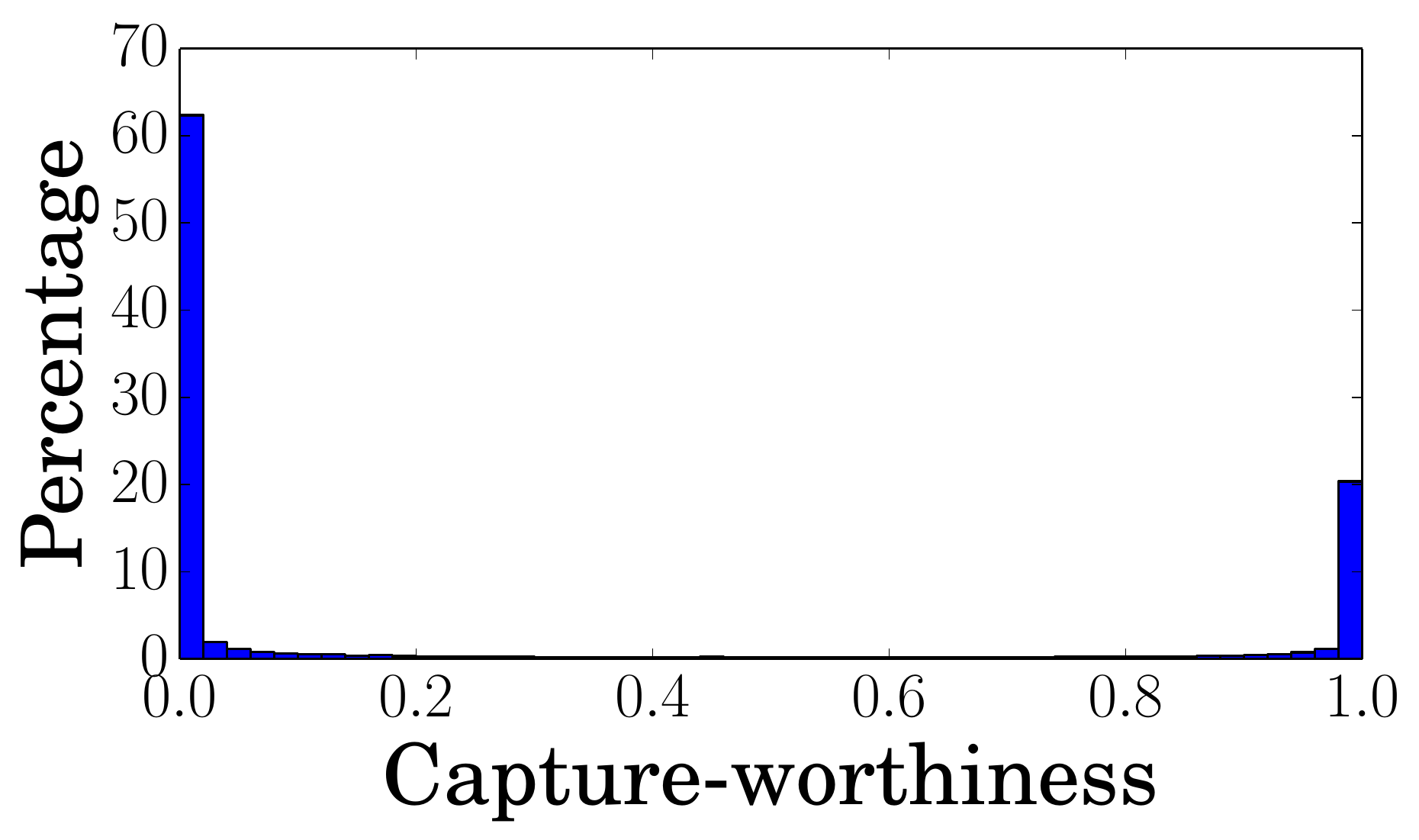}
        \vspace{-16pt}
        \caption{Value distribution.}
        \label{fig:naturalness_scores}
    \end{subfigure}
    \begin{subfigure}[t]{0.3\textwidth}
        \includegraphics[width=\textwidth]{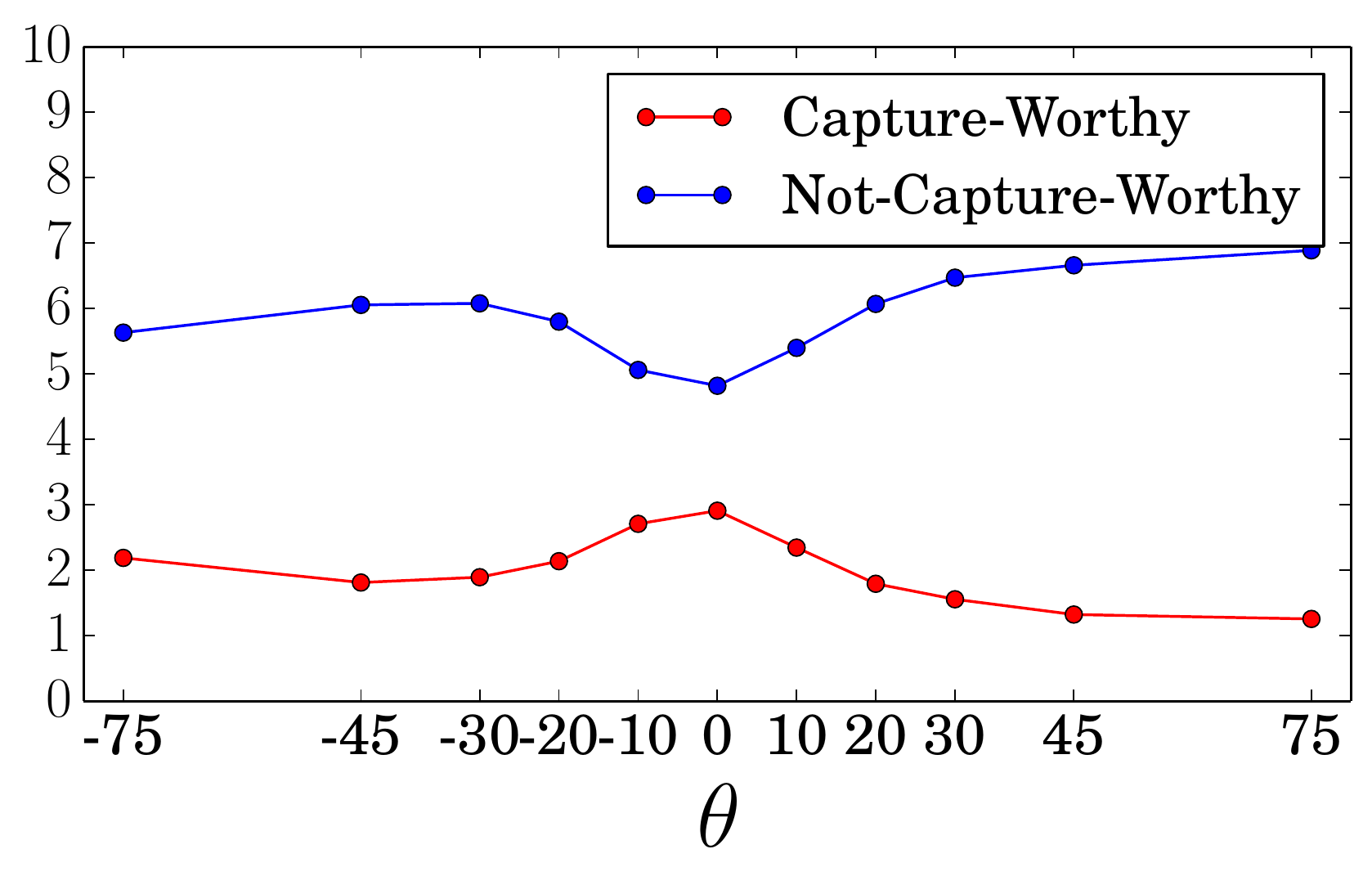}
        \vspace{-16pt}
        \caption{wrt latitude.}
        \label{fig:naturalness_latitude}
    \end{subfigure}
    \begin{subfigure}[t]{0.3\textwidth}
        \includegraphics[width=\textwidth]{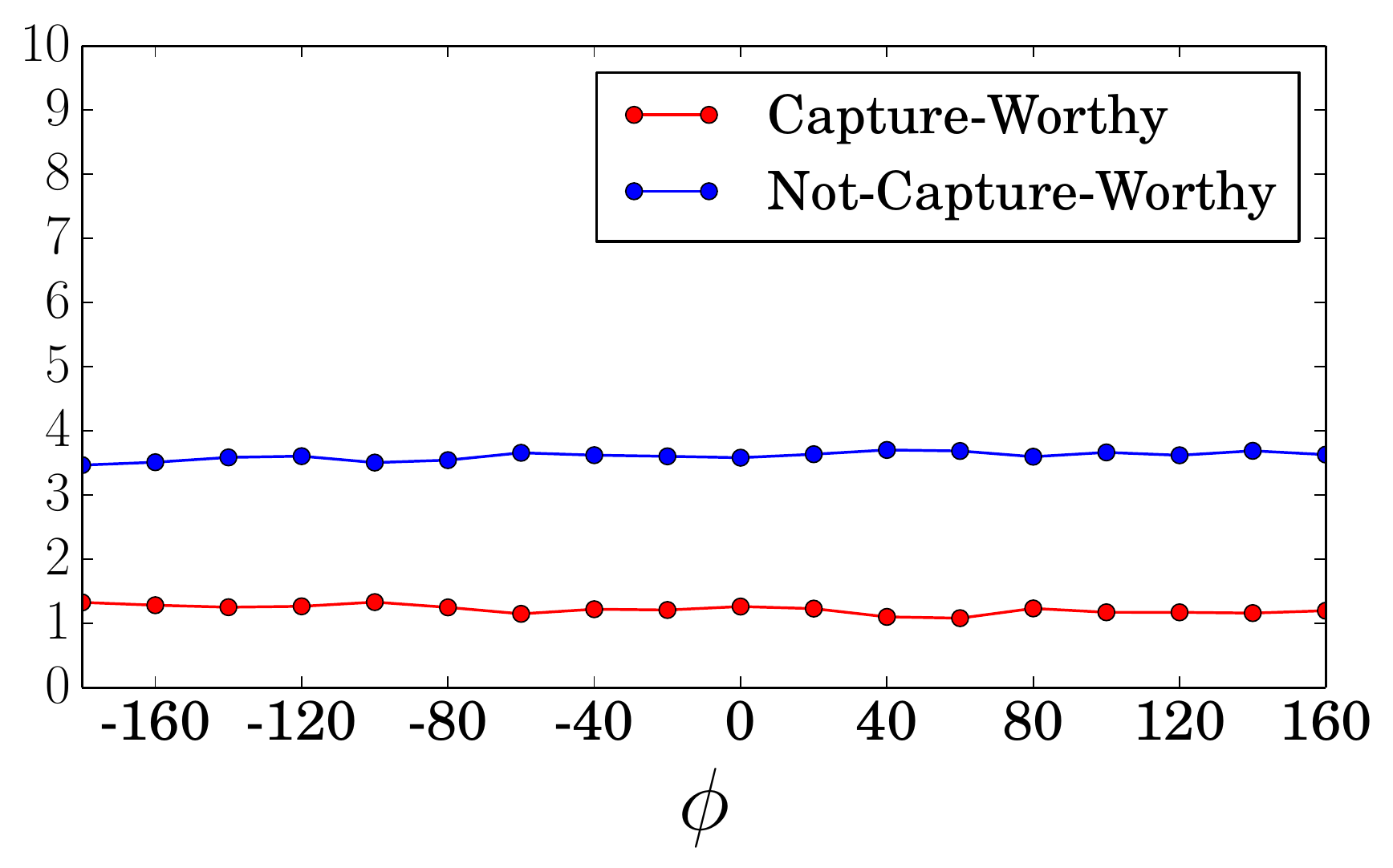}
        \vspace{-16pt}
        \caption{wrt longitude.}
        \label{fig:naturalness_longitude}
    \end{subfigure}
    \vspace{-8pt}
    \caption{Capture-worthiness score distribution analysis. Some glimpses
    indeed look ``natural,'' i.e.~having capture-worthy score $\ge 0.95$. Also,
    the score demonstrates weak eye-level preference but no center preference.}
    \label{fig:distributions}
    \vspace{-18pt}
\end{figure*}

In this section, we perform further analysis of the first stage of our
pipeline, discriminative capture-worthiness assessment.  We first show the
score distribution in Fig~\ref{fig:naturalness_scores}. It turns out to be
almost binary, indicating there is a clear distinction between ``natural'' and
``un-natural'' content and the problem of whether a NFOV video looks natural is
well defined in the feature space.  The distribution also shows that there are
some ST-glimpses indeed look natural, and using them as the basis of AutoCam is
reasonable.

We next show the score distribution with respect to latitude in
Fig~\ref{fig:naturalness_latitude} and longitude in
Fig~\ref{fig:naturalness_longitude}.  We define ``capture-worthy'' glimpses as
those with the score $\ge 0.95$ and ``non-capture-worthy''
as those with score $\le 0.05$. The y-axis is the percentage of glimpses with
respect to all candidate glimpses.

The capture-worthiness score does not have center preference but does have a
weak eye-level preference.  This difference is caused by the nature of the two
preferences.  The center preference is induced by the camera recorders that try
to align content of interest in $360\degree$ videos to the center and is only
present in some videos. On the other hand, the eye-level prior is induced by
video framing and is universal to all videos.  This is consistent with the
experiment results in Sec~4.2 where \textsc{Eye-level} performs better,
suggesting it is more universal and therefore stronger.  The preference is
nonetheless weak and ST-glimpses not at eye-level still have a good chance to
be natural.

\vspace{-12pt}
\section{HumanEdit Interface}
\label{sec:appendix_interface}

\vspace{-8pt}

In this section, we illustrate the HumanEdit interface by breaking down all the
components. \textbf{Please see the project webpage for the interface in action.}

\begin{figure}[t]
    \centering
    \vspace{-12pt}
    \includegraphics[width=\textwidth]{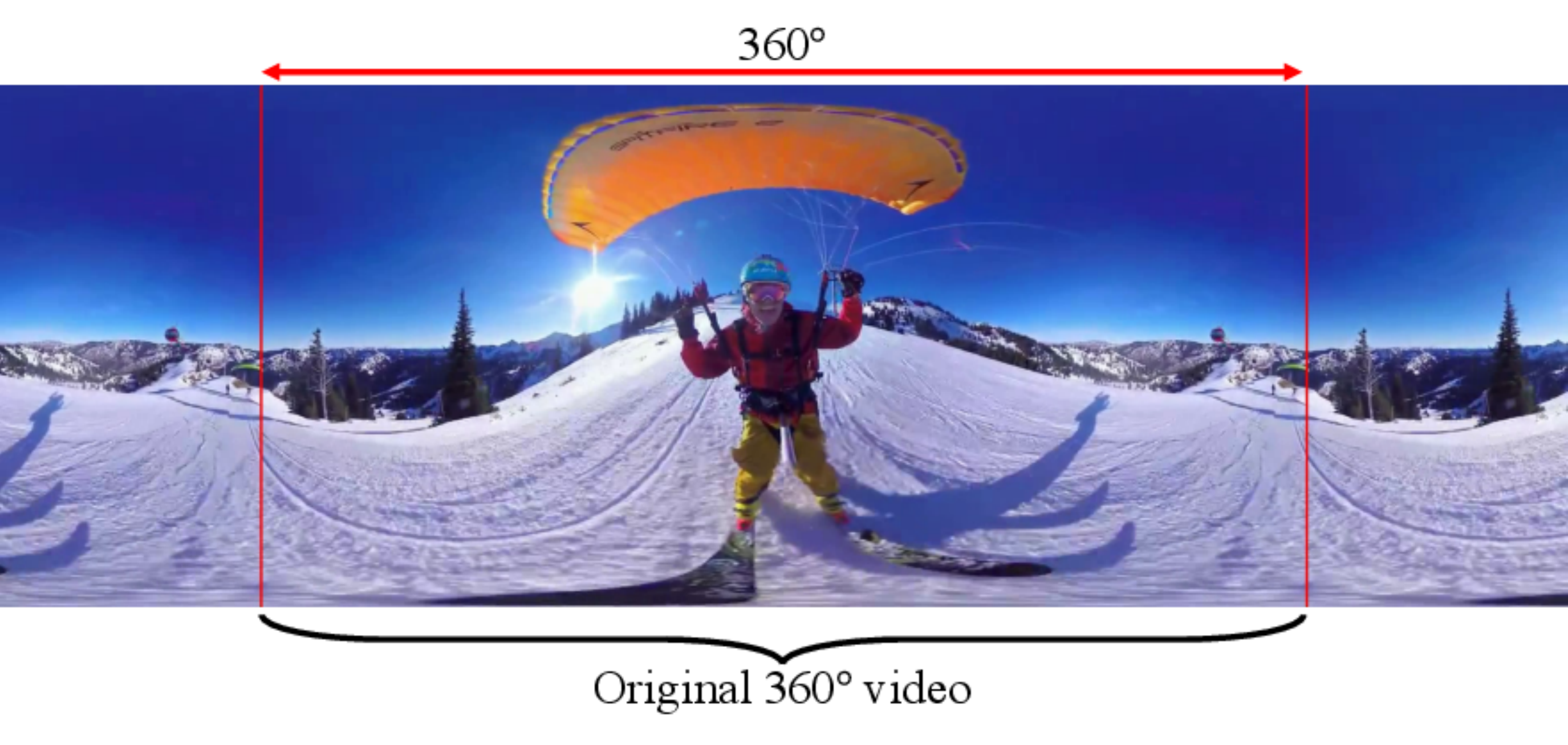}
    \vspace{-24pt}
    \caption{We display the entire $360\degree$ video in equirectangular
    projection.}
    \label{fig:panorama_video}
    \vspace{-18pt}
\end{figure}

We display the entire $360\degree$ video in equirectangular projection so the
annotator can see all video content at once. See Fig.~\ref{fig:panorama_video}.
While the goal is to capture FOV video by the virtual NFOV camera, content
outside the camera FOV provides important information for directing the camera.
This is similar to the situation when a videographer is capturing a video,
where the videographer may look at the surrounding environment without moving
the camera.

\begin{figure}[H]
    \centering
    \vspace{-16pt}
    \includegraphics[width=\textwidth]{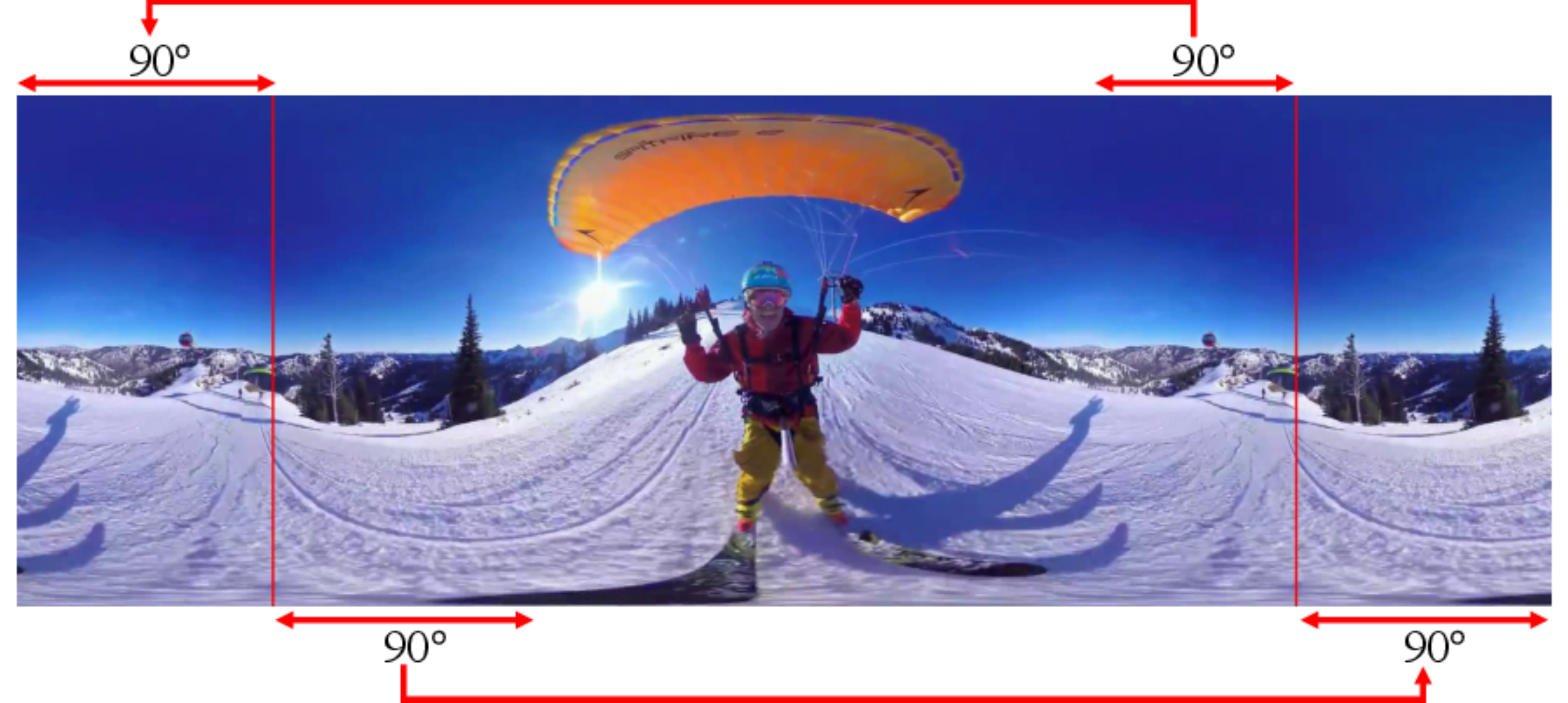}
    \vspace{-16pt}
    \caption{We extend the video horizontally by $90\degree$ on both side.}
    \label{fig:extend_fov}
    \vspace{-16pt}
\end{figure}

If we display only $360\degree$ horizontally, the $\phi = \pm 180\degree$
boundaries will appear on the opposite side of the screen and be discontinous,
even though they correspond to the same direction physically. This
discontinuity may cause difficulty for perception and annotation. To remedy
this problem, we extend the video by $90\degree$ on both left and right
(Fig.~\ref{fig:extend_fov}) so $\phi \pm \Delta\phi$ are adjacent to $\phi$ on
the screen $\forall \phi$. The $360\degree + 2\times 90\degree$ video will span
the width of the screen.

\begin{figure}[t]
    \centering
    \vspace{-4pt}
    \includegraphics[width=\textwidth]{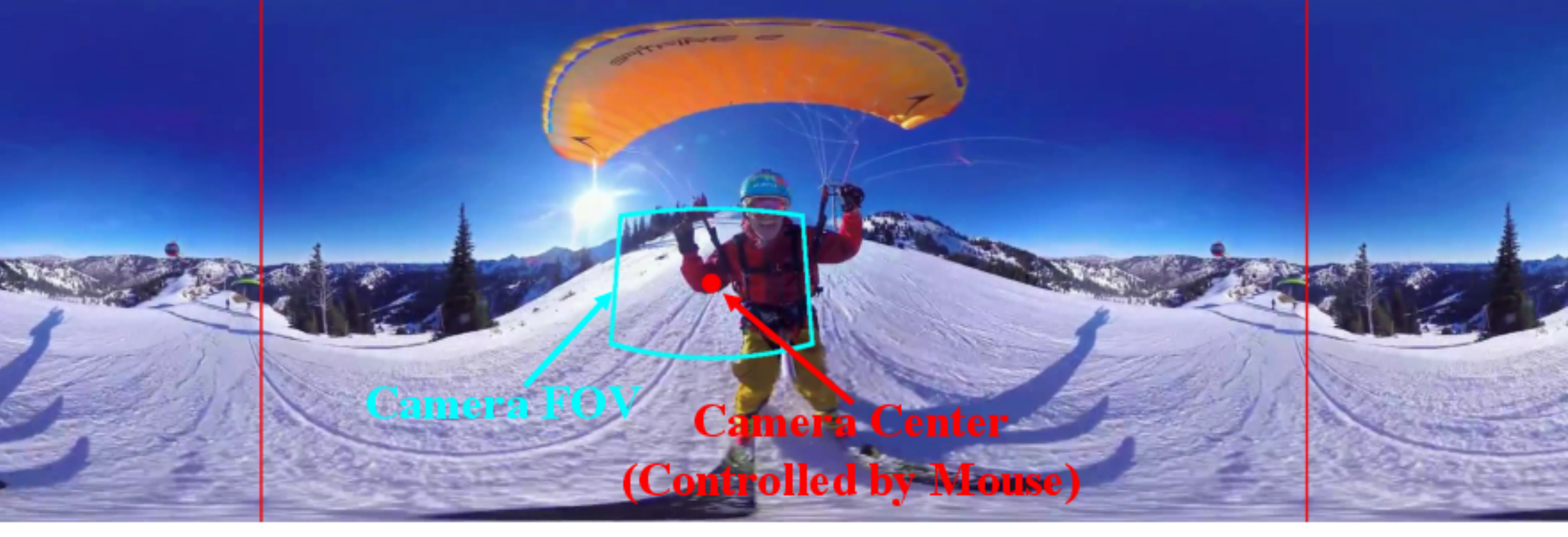}
    \vspace{-16pt}
    \caption{The annotator controls the camera using the mouse. The
    camera FOV is backprojected onto the $360\degree$ video during annotation
    to show what it captures.}
    \label{fig:backproject_fov}
    \vspace{-16pt}
\end{figure}

The annotators are asked to control the virtual camera direction within the
video using mouse while the video is playing, and we record the mouse location
throughout the video as HumanEdit. The camera center, i.e.~mouse location, is
highlighted by a red circle. To help the annotator understand what content does
the virtual camera really captures, we backproject the camera FOV on the video
(in cyan) in real time. See Fig.~\ref{fig:backproject_fov}.

\begin{figure}[H]
    \centering
    \vspace{-12pt}
    \includegraphics[width=\textwidth]{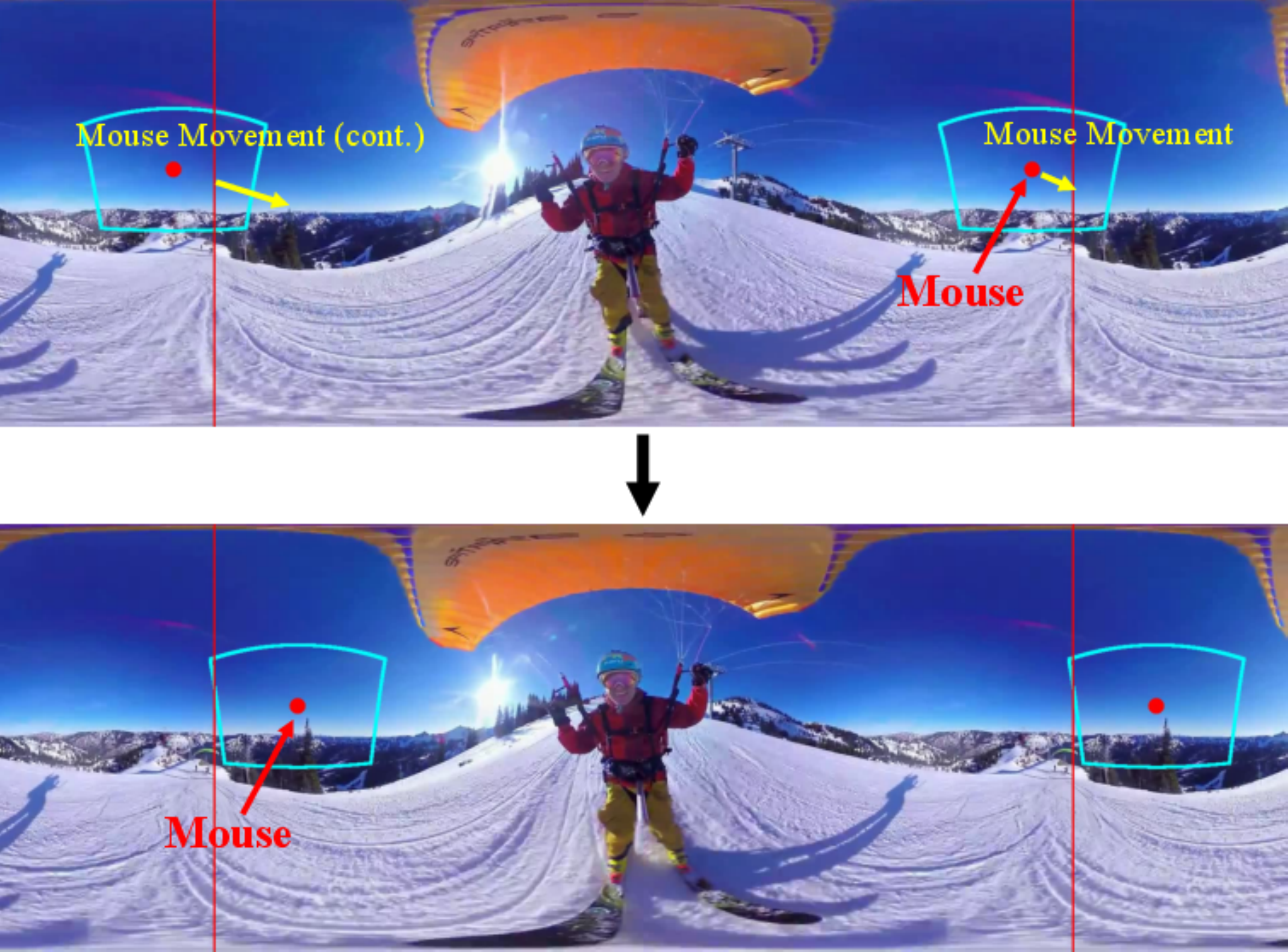}
    \vspace{-16pt}
    \caption{The mouse is restricted in the original $360\degree$ horizontally.
    If the mouse move over the edge, it will be repositioned to
    the duplicate position within $360\degree$ boundary.}
    \label{fig:mouse_movement}
    \vspace{-16pt}
\end{figure}

To simulate the smooth camera motion in the real world, we restrict the mouse
to move within the original $360\degree$ video and reposition the mouse to the
duplicate position within $360\degree$ boundaries when it passes through the
$\phi=\pm 180\degree$ edges, as shown in Fig.~\ref{fig:mouse_movement}. Because
the camera center and FOV are also duplicated, they will not disappear but
behave as if the mouse remains at the original location after it is
repositioned.

\begin{figure}[t]
  \centering
    \includegraphics[width=\textwidth]{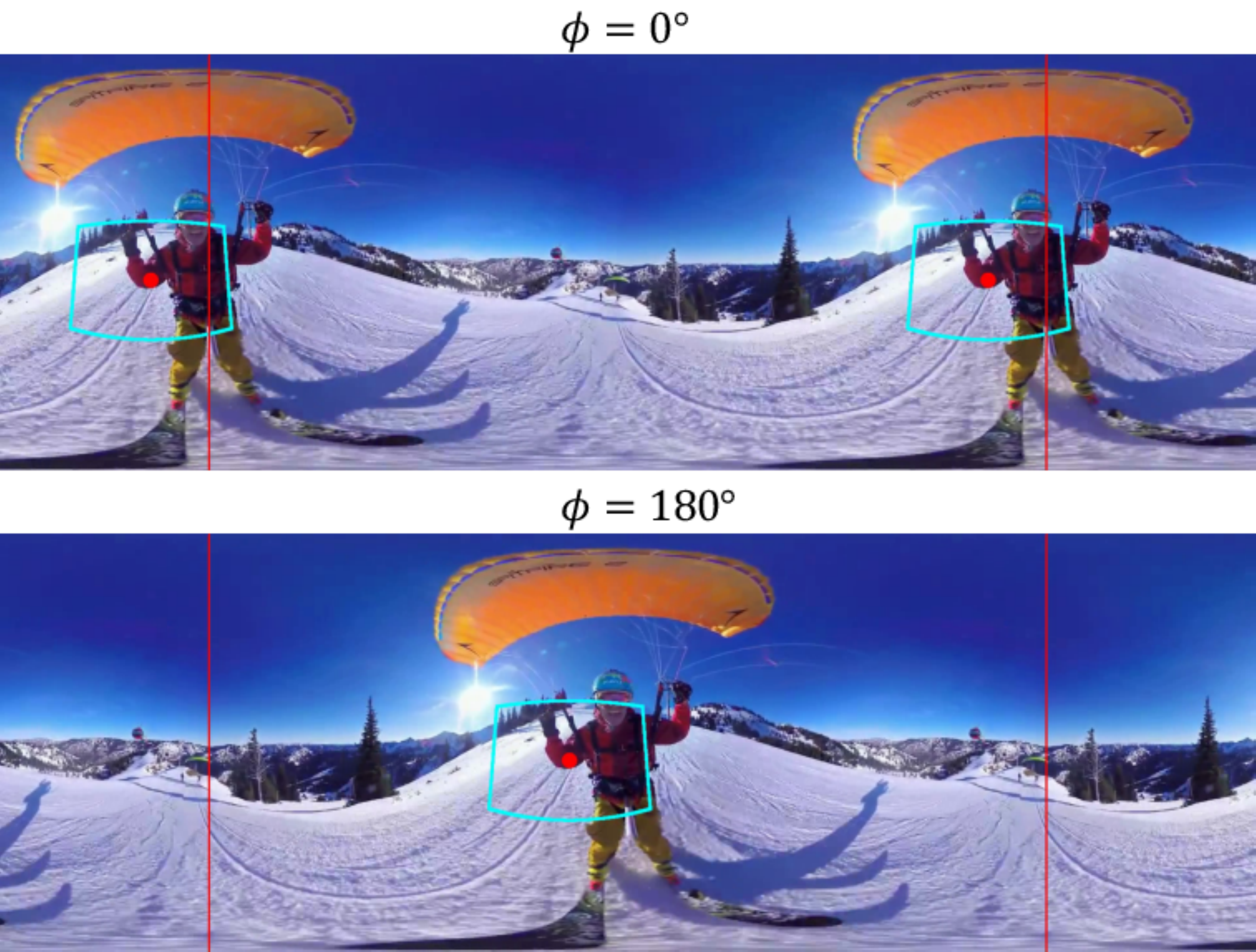}
    \vspace{-16pt}
    \caption{The annotator can select $\phi$ that corresponds to the display
    center.}
    \label{fig:select_angle}
    \vspace{-16pt}
\end{figure}

We also ask the annotator to choose the longitude for the center of display
before the video starts playing.  See Fig.~\ref{fig:select_angle}. This allows
the annotator to place the content of interest in the middle of screen and
makes annotation easier.

\vspace{-8pt}
\section{Generating Multiple Trajectories}
\label{sec:appendix_multitraj}
\vspace{-8pt}

Because there may be multiple valid trajectories in each $360\degree$ video,
instead of considering a single output, we generate $K=20$ different
trajectories by each method and consider the joint performance of these top-$K$
trajectories. In this section, we describe how to generate multiple outputs in
detail.

\begin{itemize}
    \item \textsc{AutoCam}---We use the algorithm described in the main text to
        generate one optimal trajectory for every glimpse location in the last
        frame, i.e. $\Omega_{T,\theta,\phi}\;\forall\;(\theta,\phi)$. We then
        select $K$ trajectories with maximum accumulated capture-worthiness
        scores. Note the accumulated score in intermediate frames can be
        reused, so all these trajectories can be constructed in one pass of the
        dynamic programming algorithm. The same method is used to generate $K$
        trajectories for \textsc{Saliency}.
    \item \textsc{AutoCam w/o Stitching}---This is a stochastic method that
        samples a ST-glimpse at each time step independently of others. The
        probability distribution of ST-glimpses within each frame is obtained
        by taking the softmax function on the capture-worthiness scores. We
        sample the glimpses $K$ times independently to generate $K$
        trajectories.
    \item \textsc{Center}---This is also a stochastic method where the
        trajectory starts at $\theta=0, \phi=0$ and then performs random motion
        in the following time steps. Similar to \textsc{AutoCam w/o Stitching},
        we generate multiple trajectories by performing $K$ independent
        samples.
    \item \textsc{Eye-level}---We generate trajectories with static ST-glimpse
        location $\Omega_{\theta,\phi}$ in every frame where $\theta =
        0\degree$ and $\phi \in
        \{0\degree,20\degree,40\degree,\ldots,340\degree\}$. The $\phi$ samples
        are the same as our ST-glimpses and results in $K=\frac{360}{20}=18$
        trajectories instead of $20$.
\end{itemize}

In the \textbf{Distinguishablility} and \textbf{Transferability} metrics in
HumanCam-based evaluation, the $K$ trajectories are treated as $K$ independent
samples for classification. In \textbf{HumanCam-likeness} metric, these $K$
trajectories are ranked jointly, and the average/median ranking is used to
evaluate the method on a particular $360\degree$ video. Similar to
\textbf{HumanCam-likeness}, the $K$ trajectories are evaluated independently in
HumanEdit-based evaluation and their average similarity/overlap is used as the
performance measurement on the $360\degree$ video.

\end{document}